\documentclass[11pt,a4paper]{article}
\usepackage[hyperref]{acl2021}
\usepackage{times}
\usepackage{latexsym}

\usepackage{microtype}

\aclfinalcopy % Uncomment this line for the final submission
\def\aclpaperid{483} %  Enter the acl Paper ID here

\usepackage[utf8]{inputenc}
\usepackage[T1]{fontenc}
\usepackage{amsfonts}

\usepackage{microtype}
\usepackage{graphicx}
\usepackage{subfigure}
\usepackage{booktabs}
\usepackage{enumitem}
\usepackage{amsmath}
\usepackage{url}
\usepackage{tabularx}

\usepackage{verbatim} 

\usepackage{multirow}
\usepackage{color}

\definecolor{ourlightblue}{HTML}{E0ECF7}
\definecolor{ourdarkblue}{HTML}{092E6B}
\definecolor{msgrblue}{HTML}{4889f4}
\definecolor{msgrgray}{HTML}{e1e1e7}

\definecolor{botc}{rgb}{0.458, 0.488, 0.978}

\definecolor{humanc}{rgb}{0.8, 0.8, 0.8}

\definecolor{light-gray}{gray}{0.90}
\definecolor{dark-gray}{gray}{0.30}

\definecolor{aszlam}{rgb}{1.0, 0.0, .6}

\title{Internet-Augmented Dialogue Generation}

\author{ Mojtaba Komeili \quad Kurt Shuster \quad Jason Weston\\
\\
 Facebook AI Research
}

\date{}

\begin{document}
\maketitle

\begin{abstract}
The largest store of continually updating knowledge on our planet can be accessed via internet search. In this work we study giving access to this information to conversational agents.
Large language models, even though they store an impressive amount of knowledge within their weights, are known to hallucinate facts when generating dialogue \cite{shuster2021retrieval}; moreover, those facts are frozen in  time at the point of model training. In contrast, we propose an approach that learns to generate an internet search query based on the context, and then conditions on the search results to finally generate a response, a method that can employ up-to-the-minute relevant information.
We train and evaluate such models on a newly
collected dataset of human-human conversations whereby one of the speakers is given access to internet search during knowledge-driven discussions in order to ground their responses.
%We use our  newly collected task to both train 
%and evaluate a broad class of techniques, comparing our search query-based approach to retrieval-based and non-augmented models.
We find that search-query based access of the internet in conversation provides superior performance %in both automatic and human evaluation metrics 
compared to existing approaches that 
either use no augmentation 
or FAISS-based retrieval \cite{lewis2020retrieval}. %as in previous approaches. 
\end{abstract}

\begin{figure}[t!]
  %   \centering
  \includegraphics[width=7.9cm]{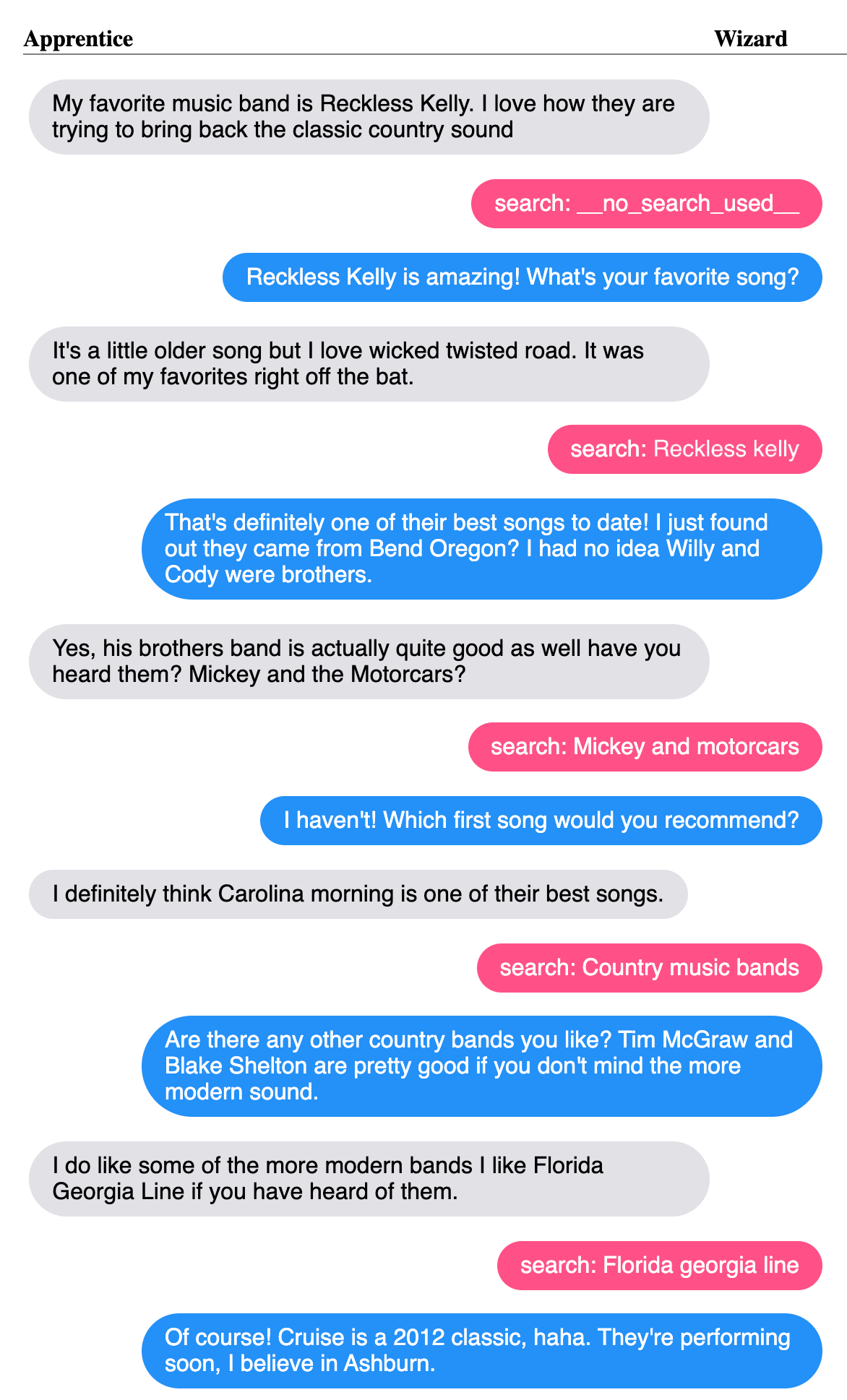}
    \caption{Example human-human conversation from the Wizard of the Internet training set. The role of the Wizard on the right-hand side involves performing internet searches, and then writing appropriate responses to the Apprentice given the viewed web documents (not shown).}
    \label{fig:train_example}
\end{figure}

\section{Introduction}

Open-domain dialogue, which involves chat about any topic, rather than a specific goal-directed topic,  is commonly studied by training large language models \cite{adiwardana2020meena,zhang2019dialogpt,roller2020recipes}. These models are trained either in a encoder-decoder or decoder only setting on large datasets of human-human conversations, and any knowledge obtained during training is stored in the weights of the model. Such static language modeling fails to take into account the dynamic state of the world, where new information is coming in by the day -- or even by the minute -- as the knowledge in static models is gleaned from the point in time when the dataset was collected, and then frozen into the model that is trained; see \citep{lazaridou2021pitfalls} for criticisms of this approach. Further, static language models are known to {\em hallucinate}, that is they  generate plausible looking statements that are factually incorrect, which can be interpreted as a form of lossy compression  when employing training to encode that knowledge within the weights of a neural network; see \citep{shuster2021retrieval} for an in-depth study.

In this work we study generative models that are instead capable of accessing the vast knowledge of the internet dynamically in order to inform their responses. Utilizing encoder-decoder architectures, we consider models that, given a dialogue context, first generate a search query. 
 The queries are then used to retrieve relevant knowledge that is prepended to the conversational history, which is encoded using the Fusion-in-Decoder method \cite{izacard2020leveraging}. Taking into account this encoded knowledge, a response is finally generated using the decoder.
  This ability to access the internet means the model is always up-to-date, unlike existing models that only know about facts in their fixed training set. Our model, in contrast, can potentially make use of the latest sports scores, movies or TV shows that were just released, the latest reviews, and so forth --  amongst the countless other topics available on the internet. 

In order to train and evaluate such models, we collect a new crowdsourced English dataset involving human-human conversations, where one of the workers plays the role of a ``wizard'' who conducts internet searches in order to inform their responses during knowledge-grounded conversations. We show that internet-augmented models trained to replace the human wizard outperform conventional non-augmented generation models 
on this task as measured by automatic metrics as well as human evaluations. We study in detail different ways of designing internet-augmentation and show which methods perform best, providing analysis of how the
 methods work, and in which conditions. We make our final models, and the new task we have collected, publicly available and open source\footnote{\url{http://parl.ai/projects/sea}}.

\section{Related Work}

The majority of work on dialogue generation has focused on training 
 on natural or crowdsourced data where the task is, given a dialogue context (history), to generate the next response. Datasets such as 
pushshift.io Reddit \cite{baumgartner2020pushshift}, PersonaChat \cite{zhang2018personalizing} or Empathetic Dialogues \cite{rashkin2019empathetic} (see \citet{huang2020challenges} for a review) are typically employed to train the weights of a Transformer encoder-decoder. This is the standard approach in state-of-the-art chatbots such as Meena \cite{adiwardana2020meena} or BlenderBot \cite{roller2020recipes}. 
Such models do not augment their generations with access to external knowledge,  instead relying on facts originally provided in the training datasets themselves  being stored into the weights of the model. 

A growing area of research is that of augmenting generative models with external knowledge. Earlier works such as Memory Networks \cite{weston2014memory} and DrQA \cite{chen2017reading} utilized TFIDF-based retrieval over documents to provide additional input to neural models for the task of question answering, following the well studied area of non-neural methods that use retrieval for QA \cite{voorhees2001trec}.
 More recently, the RAG (Retrieval-Augmented Generation) \cite{lewis2020retrieval} and FiD (Fusion-in-Decoder) \cite{izacard2020leveraging} models developed these ideas further, using a neural retriever as well, with superior results.
% in LMs
Retrieval-augmentation is also studied in  the area of  language modeling, where it is used for pre-training \cite{guu2020realm}, and as a memory \cite{yogatama2021adaptive}, especially using $k$-nearest neighbor-based cache models \cite{kh2020nearest,Khandelwal2020Generalization,grave2016improving,merity2016pointer}.

% dialogue and knowledge
In dialogue, knowledge grounding is becoming more popular an area, with several datasets developed to study it \cite{zhou2018dataset,dinan2018wizard,ghazvininejad2018knowledge,gopalakrishnan2019topical,galetzka2020corpus}.  Some of these such as Topical-Chat \cite{gopalakrishnan2019topical} and CMU\_Dog \cite{zhou2018dataset} are constructed given a gold passage of knowledge, and the task analyzes whether the model can use this knowledge in dialogue.
Other works \cite{Zhao_2020,Kim2020Sequential,Bruyn2020BARTFK} study whether knowledge selection is possible from a (small) set of knowledge.
However, a retrieval step (or search engine) is not used, as we consider here.

Perhaps the closest to our work is the Wizard of Wikipedia task \cite{dinan2018wizard} which involves conversations grounded in Wikipedia, using a TFIDF retrieval model to find relevant knowledge from that database.
Our work can be seen as a much richer task, covering all of the information that is publicly available on the internet and hence a more diverse range of conversational topics rather than just Wikipedia, while allowing human wizards to search for relevant knowledge themselves. Moreover, we consider  sophisticated neural-in-the-loop retrieval mechanisms and real search engines. \citet{shuster2021retrieval} studied neural-retriever-in-the-loop methods on this dataset.

Some other related work of note is that of using search engines for machine translation (rather than dialogue, as we do here), which was shown to provide good results
\cite{gu2018search}.
Finally, rather than applying search engines for a downstream task, sophisticated machine learning has also been applied directly to improve the search task itself, for example using reinforcement learning \cite{nogueira2017task}.

\section{Internet-Augmented Generation}

We consider two ways to access the webpages from the internet: 
(i) using a cached set of pages that are stored in a distributed approximate nearest-neighbor database, FAISS \cite{johnson2019billion}, or (ii) using an Internet Search Engine directly to retrieve pages.  For the FAISS-based methods, there are a number of possible variants that we consider, which we will describe first.

\subsection{FAISS-based methods}

In our experiments, the FAISS-based methods share the same core setup. First, we store and utilize the Common Crawl dump of the internet from \citet{wenzek2019ccnet}\footnote{We use the November 2020 dump, head only, consisting of $\sim$109M English webpages. Each document is split into 100-word chunks, giving 250M passages to index in FAISS. We also consider  the dump of Wikipedia from \cite{karpukhin2020dense} in this work.} in a FAISS database, with keys that are dense vectors. The retrieval system 
uses a DPR (Dense Passage Retrieval) \cite{karpukhin2020dense} Transformer-based model which scores document-context pairs in order to rank them based on their match using a bi-encoder framework, where the base DPR model is pre-trained on QA data pairs.  We use the pre-trained DPR model from the KILT Benchmark \cite{petroni2020kilt}.  The documents (webpages) are encoded using DPR into dense vectors and these are stored in the FAISS index. During dialogue-based retrieval, the dialogue context is also encoded by DPR into a dense vector and FAISS approximate nearest-neighbor lookup is performed, where the top $N$ documents are returned. We then consider several recent neural methods for utilizing this  retrieval mechanism in various ways.

\paragraph{RAG (Retrieval Augmented Generation)}
RAG \cite{lewis2020retrieval} is an approach which consists of two components which are trained end-to-end: (i) the neural-in-the-loop retrieval system; and (ii) an encoder-decoder for generating final responses given the results of the retrieval.
Using DPR, the top $N$ documents are returned as described above, and in the RAG-Token model (just called RAG in the rest of the paper) each in turn is encoded along with the context for each token, and the most likely sequence is generated from the set. During backpropagation training steps, the DPR context encoder is also tuned to perform well at FAISS retrieval, but the document encodings are held fixed. This approach has been shown to optimize both retrieval and generation jointly, improving results. 

\if 0
OLD:
\paragraph{RAG (Retrieval Augmented Generation)}
RAG \cite{lewis2020retrieval} is an approach which consists of two components which are trained end-to-end: (i) a neural-in-the-loop retrieval system; and (ii) an encoder-decoder for generating final responses given the results of the retrieval.
%The retrieval system
%uses a DPR (Dense Passage Retrieval) \cite{karpukhin2020dense} Transformer-based model
%which scores document-context pairs in order to rank them based on their match using a bi-encoder framework, where the base DPR model is pre-trained on QA data pairs. 
The documents are encoded using DPR into dense vectors and these are stored in the FAISS index. During retrieval, the context is first encoded by DPR also into a dense vector and FAISS approximate nearest-neighbor lookup is performed. The top $N$ documents are returned, and in the RAG-sequence model each in turn is encoded along with the context, and the most likely sequence is generated from the set. During backpropagation training steps, the DPR context encoder is also tuned to perform well at FAISS retrieval, but the document encodings are held fixed. This approach has been shown to optimize both retrieval and generation jointly. In our work, we simply use this existing approach but apply it to dialogue data, and use our Common Crawl dump as the database, start with the DPR model from \cite{petroni2020kilt}, and fine-tune the whole system it on our knowledge-from-the-internet-grounded training data (\autoref{sec:WoI}).
\fi

\paragraph{FiD (Fusion in Decoder)}
A related, but perhaps simpler, method is that of FiD \cite{izacard2020leveraging}. In this case, the pre-trained retriever is used, i.e. DPR with FAISS, and then
each of the top $N$ documents returned is prepended to the context and encoded separately by the encoder, and finally all the results are concatenated. The decoder then attends to these encodings to produce a final response, so all ``fusion'' happens in the decoding stage. This relatively simple method was shown to outperform RAG in some cases.

\paragraph{FiD-RAG}
The FiD approach works well, but there is no end-to-end training of the retriever in that case, and so it relies completely on being pre-trained well, as opposed to RAG which tunes the retrieval for generation. FiD-RAG, proposed in \cite{shuster2021retrieval} combines the two methods. First the retriever is trained in a RAG setup, and then FiD is used with that retriever. This was shown to give superior results to both RAG and FiD on dialogue tasks.

\paragraph{FAISS + Search Query-based Retrieval}

Instead of just encoding the context into a dense vector, in this approach an encoder-decoder is employed to generate a search query given the context. The search query is input into a DPR model to produce a dense vector, and is matched to documents in the FAISS index. Returned documents can then be used in the final response generation encoder-decoder as before.  Any of the existing approaches (RAG, FiD or FiD-RAG) could potentially be used to fuse the DPR and generator models. We used the standard DPR FiD setup. We will discuss how to generate the search query itself in more detail in the following subsection (\autoref{sec:sq}).

\subsection{Search Engine-Augmented Generation  (SEA)} \label{sec:sq}

The previously described FAISS-based approaches can take advantage of many existing methods developed for QA and dialogue tasks, as we saw, but have several disadvantages. First, they may be difficult to update to real-time web documents; second, there may be a limit to the number of documents storable in local FAISS deployments; and third, such methods
will not take advantage of the high quality ranking that has been finely tuned in Internet Search engines over decades of use. We thus
consider using Internet search engines directly.

\paragraph{Method}
Our proposed method consists of two components:
\begin{itemize}
    \item A search query generator: an encoder-decoder Transformer that takes in the dialogue context as input, and generates a search query. This is given to the black-box search engine API, and $N$ documents are returned.
    \item A FiD-style encoder-decoder model that encodes each document individually, concatenates them to the dialogue context encoding, and then finally generates the next response. 
\end{itemize}

We can train each of these modules separately if we have supervised data available for both tasks, the first module requiring (context, search query) pairs, and the second module requiring (context, response) pairs. As we will see, the data we collect in this work (detailed in \autoref{sec:WoI})  fulfills both of these requirements. 

For FiD, we try two methods: (i) Conventional FiD whereby we use the returned search results from using our trained search query generator in order to build the relevant document contexts for the FiD training set; %In this case we use the search query generator as input to the search engine for that goal. 
(ii) FiD-Gold: as we will have available human-written search queries for the training set, and their corresponding search results, we can use these gold results to build training document contexts instead. 
Although these might not look like the queries and hence results predicted at test time, they are more likely to contain the knowledge used in generating the training set responses, thus a clearer grounding may be apparent for the model to learn correspondences.

\paragraph{Search Engine} The search engine is a black box in this system, and could potentially be swapped out for any method.  In our numerical experiments we use the Bing Search API  to generate a list of URLs for each query; then, 
%we look up these URLs in our same Common Crawl dump to populate a set of pages for that query.
we use these URLs as keys to find their page content from a lookup table we built for our Common Crawl snapshot, in order to populate a set of pages for that query.
This makes our comparison more direct with our FAISS-based methods.
In addition, we can also consider if the URL is from English Wikipedia, in that case we can extract the page title from the URL and look up its corresponding page inside the dump of Wikipedia.

\subsection{Knowledge Response Regularization}
\label{sec:krr}

It has been observed before that large language models,  when augmented with retrieval, 
have trouble with choosing between copying knowledge remembered within their weights and
knowledge provided in retrieved documents \cite{shuster2021retrieval}.
Here, we propose a general regularization method to more finely control this mechanism:
when training, we multi-task between the original response generation task and a new task
which consists of generating the selected knowledge from retrieved documents indicated by human annotators\footnote{We note that this technique is similar to the one used in retrieve and refine architectures \cite{roller2020recipes}.}. The second task can be seen as a regularizer that encourages the use of retrieved documents, as the easiest way for the model to do well on that task is to attend and copy to the document where that text already exists. Then, by changing the mixing parameter between the two tasks, the intent is to achieve a smooth control between encouraging copying from retrieved documents, or not.

\section{Wizard of the Internet Task} \label{sec:WoI}

\begin{table*}
    \centering
    \small
    \begin{tabular}{lrrrr}
{\bf Wizard of The Internet Task} &  Train & Valid & Test & Total \\
\toprule
Number of Dialogues         & 8,614  & 516 & 503 & 9,633\\ 
Number of Utterances        &  82,952 & 5,781 & 4,932 & 93,665 \\
Average Utterance Length    & 18.67 & 22.9 & 21.5 & 19.1 \\
Average Utterances per Dialogue & 9.6 & 11.2 & 9.8 & 9.7 \\
Number of Searches & 42,306 & 3,306 & 2,763 & 48,375 \\
Number of unique URLs selected & 26,192 & 2,087 & 1,973 & 29,500 \\
Number of unique Domains selected & 10,895 & 1,256 & 1,256 & 11,963 \\
%\midrule
%{\bf Common Crawl}   & \multicolumn{3}{c}{672M pages ~~~~ 12.5B sentences} \\
\bottomrule
\end{tabular}
\caption{{\bf Wizard of the Internet (WizInt) Dataset Statistics.} 
\label{tab:stats}
}
\end{table*}

\begin{figure}
    \centering
  \includegraphics[width=7.8cm]{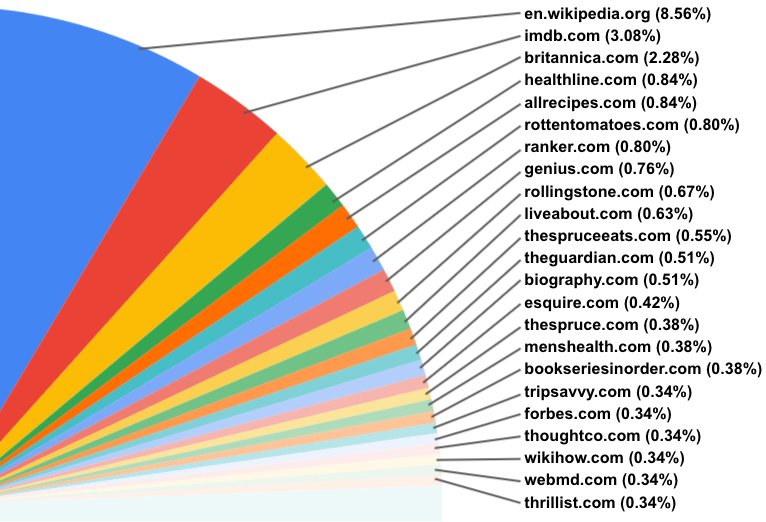}
    \caption{Breakdown of most common domains used during search by the wizard in our newly collected dataset (validation set breakdown). Shown is the most common 24.41\%, there is a long tail of 1233 other domains across the whole validation set.} %, and 2087 unique URLs in total.}
    \label{fig:pie}
\end{figure}

In order to both train and evaluate  generative models that can use search engines in-the-loop, we design, collect and release a dataset for this purpose. The overall setup involves pairing crowdworkers that are instructed to have a conversation together. One plays the role of the {\em wizard}, who has access to a search engine during conversation, while the other, the {\em apprentice}, does not. The {\em apprentice} however has an assigned persona that describes their interests. The purpose of the exchange is to have an ``in-depth conversation about [those] assigned interests''. This mirrors conversations we expect to be more prevalent between a human and a bot: the conversations are more likely to be centered around the human's interests than the bot's, and the bot is the one that is going to be using the search engine to ground their knowledge. Hence, when we train or evaluate on this task, a given model will replace the role of the {wizard}. %, who engages with a human ({\em apprentice}).

\paragraph{Apprentice Persona} We show the apprentice several possible persona choices for the character that they are going to play, and let them choose one, e.g. ``I love tennis, Rafael Nadal is my favorite player.''. The intent here is that they can choose a topic they are both more interested in themselves to talk about and also have enough knowledge of so that they can conduct a reasonable conversation. The choices we show are themselves mined from the interests provided in the existing Persona-Chat dataset \cite{zhang2018personalizing} and the topics given in the existing Topical-Chat dataset \cite{gopalakrishnan2019topical}.
More details of the choices we give are provided in \autoref{app:task}.

\paragraph{Wizard Active and Passive Openings} We randomize which speaker takes their turn first. If the wizard speaks first, we encourage them to start with an opening that addresses the apprentice's interests. For example, if they know their partner is interested in tennis, they could search for the latest tennis news, and open with an interesting point based on that knowledge. If the apprentice goes first, their goal is to converse with the wizard more based on their own interests, e.g. in this same case they could talk about tennis in detail.

\paragraph{Wizard Search} At each turn, the wizard can enter free text search terms in a left-hand panel (with the main conversation panel on the right) much like in a conventional search engine. The top few results are shown in the left panel, below the search query\footnote{We run two searches, one with the given query, and one with the query terms plus the word ``news'' (with the news results shown as the top two knowledge candidates), in order to encourage topical discussion.}. For each document the titles are shown for space reasons, and each document is expandable.  If the wizard finds one or more search results useful for their response, they can click on the sentences they find relevant, and then enter their conversational response in the right-hand panel. They are also free to try another search query if they did not find their first results appropriate, or else can enter a conversational response and choose to ignore the search results entirely.

\paragraph{Full System} Each crowdworker has to pass an onboarding task to be able to be part of the main data collection task, and pass some automatic checks (average response length, use of search). 
They are asked to play a particular role ("Create an interesting character that you want to play"), and are given instructions to avoid toxic or biased language. We randomly assign for any given crowdworker a fixed choice of either wizard or apprentice for all of their data collection, otherwise we found that switching role introduced lower quality conversations, probably due to confusion between the different goals and instructions per role. After pairing, we collect between 5-6 turns (10-12 utterances) for each conversation. We ask workers to skip initial greeting messages, as these bring little extra value to the task. Screenshots of the crowdworker task can be seen in \autoref{fig:mturk_screenshots} in the appendix. Example collected dialogues are shown in 
\autoref{fig:train_example} and \autoref{fig:train_example2}.

\begin{table*}[t!]
    \centering
    \small
    \begin{tabular}{llrrrrrr}
            	          & Pre-train  & \multicolumn{3}{c}{WizInt Validation} \\
        Model	          & Model    &  PPL & F1 & KF1\\
\hline
Transformer (no knowledge) & BlenderBot 2.7B & 9.9 & 18.0 & 6.6  \\ 
Transformer (no knowledge) & BlenderBot 400M & 13.4 & 17.3 & 6.2  \\ 
Transformer (no knowledge) & BART-Large 400M & 17.4 & 17.6 & 6.8  \\ 
Transformer (no knowledge) & T5-Large 770M & 15.9 & 17.9 & 6.5  \\ 
\hline
Transformer (gold knowledge) & BlenderBot 2.7B & 8.1 & 21.7 & 23.3  \\ 
Transformer (gold knowledge) & BlenderBot 400M & 9.2 & 22.0 & 22.8  \\ 
Transformer (gold knowledge) & BART-Large 400M& 10.6 & 25.4 & 23.1  \\ 
Transformer (gold knowledge) & T5-Large 770M & 10.1 & 25.7 & 23.5 \\
\end{tabular}
\caption{{\bf Choice of Pre-training Model.} We compare several pre-trained models fine-tuned on the WizInternet task, using either no or gold knowledge, measured on the validation set. Perplexities cannot be compared due to differing dictionaries except between BlenderBot 2.7B and 400M. 
\label{tab:pretrain}
} 
%\end{table*}
%\begin{table*}
    \centering
    \small
    \begin{tabular}{llrrr|rrr}
            	          & Training  & \multicolumn{3}{c}{WoW Validation}  & \multicolumn{3}{c}{WizInt Validation} \\
        Model	          & Data    & PPL & F1 & KF1 & PPL & F1 & KF1\\
\hline
Transformer (no knowledge)  & WoW  & 14.8 &  21.0 & 17.7 &  20.4 &	15.8 &	6.7 \\
Transformer (no knowledge)  & WizInt  & 22.4 & 16.7 & 13.1 & 17.4 & 17.6 & 6.8 \\
Transformer (no knowledge)  & WoW + WizInt & 15.4 & 20.0 & 16.3 & 17.3 & 18.0 & 6.9 \\
\hline
Transformer (gold knowledge) & WoW & 7.9 & 39.1 & 61.2 & 12.8	& 20.6	& 26.1\\  
Transformer (gold knowledge) & WizInt & 9.4 & 34.6 & 52.6 &  10.6 & 25.4 & 23.1  \\ 
Transformer (gold knowledge) & WoW + WizInt & 7.9 & 38.5 & 65.6 &  10.3 & 26.3 & 24.2  \\ 
\end{tabular}
\caption{{\bf Usage of the Wizard of Wikipedia Dataset with Multi-Tasking} using  BART-Large, measured on the validation set.}  
\label{tab:wow-wiz}
\end{table*}

\subsection{Overall Dataset}

% splits, stats.
The overall collected data consists of 9633 dialogues in total, with 82952 utterances in the training set,  and validation and test sets of 5781 and 4932 utterances, respectively. Overall statistics can be found in \autoref{tab:stats}.
We find that 84.81\% of all turns by the wizard involve search, so a large amount of knowledge grounding based on internet results is taking place. Of those, the wizard is allowed to repeat the search with different search terms if they did not find what they were looking for. When the wizard searches, we find 1.19 search queries are performed on average, so while mostly a single search is employed, a number of further knowledge searches are attempted. Wizards use the search results (indicated by selecting relevant sentences) 80.3\% of the time.

We show in \autoref{fig:pie} a breakdown of the most common domains used during search on the validation set.
We see that the domains are rather diverse, coming from all kinds of topics, and in particular that the Wikipedia domain is actually fairly small (8.56\% of queries), which is interesting because most other studies have used Wikipedia only as their knowledge resource \cite{chen2017reading,lewis2020retrieval,dinan2018wizard,shuster2021retrieval}. Our training
set spans 26192 unique selected URLS for grounding knowledge from 10895 domains, indicating a wide variety of topics and knowledge is used across all conversations.

\begin{table*}[hbt!]
    \centering
    \small
    \begin{tabular}{lllrrr}
            	          & {Knowledge} & {Knowledge} \\
        Model	          & Access Method &  Source    & PPL & F1 & KF1\\
\hline
%Repeat Last utterance & 		None   & None & \\
WoW Transformer (no knowledge)  &      None & 	None & 22.3  & 14.7 & 6.7 \\  
WizInternet Transformer (no knowledge)  &      None & None &  18.7 & 16.9 & 6.8 \\   
\hline
% WoW FiD 	& DPR+FAISS  & Wikipedia &\\
WoW FiD 	& DPR+FAISS  & Wikipedia &  23.0 & 14.7 & 7.4 \\ 
WoW FiD 	& DPR+FAISS  & CC        & 22.8 & 15.3 & 7.3 \\
WoW FiD-RAG & DPR+FAISS  & CC        & 22.3 & 15.5 & 7.2 \\
WoW Search engine FiD  & Bing Search & CC & 21.9 & 14.3 & 7.3 \\
\hline
WizInternet FiD-RAG 	& DPR+FAISS	& CC & 18.8	& 17.0 & 6.7 \\
WizInternet Search term FiD &	Search Query+FAISS	& CC &  19.0 & 16.5 & 6.7 \\
WizInternet Search engine FiD &	Bing Search	& CC & 17.7 & 16.8 & 6.9 \\
WizInternet Search engine FiD  &	Bing Search	& CC+Wikipedia & 17.7 &	16.6 & 6.7 \\
\end{tabular}
\caption{{\bf Results using Automatic Metrics}  measured on the test set. All models use BART-Large as a base. } 
\label{table:main_table}
%\end{table*}
%\begin{table*}
    \centering
    \small
    \begin{tabular}{lllrrr}
            	          & {Knowledge} & {Knowledge} & \multicolumn{3}{c}{WizInt Validation}\\ 
        Model	          & Access Method &  Source    & PPL & F1 & KF1\\
\hline
WoW Transformer (no knowledge)  &      None & 	None                & 20.4 & 15.8 & 6.7\\  % Max KF1 = 7.4
WizInternet Transformer (no knowledge)  &      None      & 	None    &  17.4 & 17.6 & 6.8\\    %WoW + WizInt Transformer (no knowledge)  &      None      & 	None    & 
\hline
%WoW RAG	Token	 & DPR+FAISS  & Wikipedia \\
%WoW RAG Sequence & DPR+FAISS  & Wikipedia\\
%WoW RAG	Token	 & DPR+FAISS  & CC \\
%WoW RAG Sequence & DPR+FAISS  & CC \\
WoW FiD 	& DPR+FAISS  & Wikipedia & 20.9 & 15.7 &  7.5 \\
WoW FiD 	& DPR+FAISS  & CC        & 20.8 & 16.4 & 7.4 \\
WoW RAG          & DPR+FAISS  & Wikipedia & 20.0 & 15.4 & 7.0 \\ % Trained with Wiki index
WoW RAG          & DPR+FAISS  & CC & 20.2 & 16.3 & 6.5 \\  % Trained with Wiki index
%WoW FiD-RAG		& DPR+FAISS  & Wikipedia \\
WoW FiD-RAG		& DPR+FAISS  & CC & 19.7 & 16.2 & 6.6 \\  % Trained with Wiki index
WoW Search term FiD & Search Query+FAISS & Wikipedia &  21.0 & 15.4 & 7.4  \\  % Trained with Wiki index. title generator
WoW Search term FiD & Search Query+FAISS  & CC & 20.8 & 16.3 & 7.2   \\  % Trained with Wiki index. title generator
WoW Search engine FiD & Bing Search & CC  & 19.9 & 15.4 & 7.5 \\  % Trained with Wiki index. title generator, (256 token length head chunk)
%WoW Search engine FiD & Bing Search & Wikipedia \\ 
\hline
WizInt RAG	    	 & DPR+FAISS  & Wikipedia               & 17.5 & 17.7 & 6.6  \\  % Trained with Wiki index
WizInt RAG  & DPR+FAISS  & CC                               & 17.8 & 17.7 & 6.7 \\ % Trained with CC index
WizInt FiD-RAG		& DPR+FAISS  & Wikipedia                & 17.1 & 18.0 & 7.0 \\  % Trained with Wiki index
WizInt FiD-RAG		& DPR+FAISS  & CC                       & 17.4 & 17.9 & 6.8 \\ % Trained with CC index
WizInt Search term FiD & Search Query+FAISS & Wikipedia     & 17.2 & 17.8 & 6.5 \\ % Trained with SQ + FAISS
WizInt Search term FiD & Search Query+FAISS  & CC           & 17.8 & 17.7 & 6.6 \\ % Trained with SQ + FAISS, (256 token head chunk, the document itself is only 100 words)
WizInt Search engine FiD-Gold & Bing Search & CC            & 17.6 & 14.1 & 7.4 \\ % Trained with gold retrieved docs, (256 token head chunk)
WizInt Search engine FiD-Gold & Bing Search & CC+Wikipedia  & 17.6 & 14.1 & 7.5 \\ % Trained with gold retrieved docs, (256 token head chunk)
WizInt Search engine FiD-Gold & Retrieved Gold  & CC+Wikipedia      & 13.9 & 20.0 & 9.6 \\ % Trained with gold retrieved docs, (100 word head chunk)
WizInt Search engine FiD & Bing Search & CC                 & 16.3 & 17.7 & 7.0\\  % Trained with Search Engine, (256 token head chunk)
WizInt Search engine FiD & Bing Search & CC+Wikipedia       & 16.4 & 17.9 & 6.9 \\ % Trained with Search Engine, (256 token head chunk)
WizInt Search engine FiD & Retrieved Gold  & CC+Wikipedia      & 13.8 &	18.1 &	7.5 \\ % Trained with Search Engine, (256 token head chunk, the document itself is only 100 words)
%WizInt Search term FiD-Gold & Search Query+FAISS & Wikipedia& \\
%WizInt Search term FiD-Gold & Search Query+FAISS  & CC      & \\
\hline
WoW+WizInt Search engine FiD  & Bing Search & CC+Wikipedia & 16.1 & 17.9 & 7.0 \\ % (256 token head chunk for CC docs)
\end{tabular}
\caption{{\bf Full Set of Retrieval and Search Augmentation Method Results} using automatic metrics measured on the validation set. All models use BART-Large as a base. } 
\label{table:full_set}
\end{table*}

\section{Experiments}

\subsection{Experiment and Evaluation Setup}

We evaluate models on our new Wizard of the Internet (WizInt) task, using its dedicated training set. We also consider the existing Wizard of Wikipedia (WoW) training resource as well, either for building baselines or for multi-tasking.
We consider fine-tuning various existing pre-trained models: 
T5 \cite{raffel2019exploring}, BART-Large \cite{lewis2019bart} and BlenderBot variants \cite{roller2020recipes}. 
For all retrieval-augmented methods we use $N=5$ returned documents. For all models, when generating responses we fix the decoding parameters to beam search (beam size 3) with a minimum sequence length of 20 and beam blocking of 3-grams within the response (but not the context), similar to choices in \cite{roller2020recipes}. 

Following \citet{shuster2021retrieval}  we report perplexity, F1 and Knowledge F1 (KF1) metrics. F1 measures the overlap between the model's response and the
human response from the dataset. KF1 instead measures  the overlap between the model's response and the knowledge on which the human grounded during
dataset collection (i.e., the sentences they clicked as relevant from the web search documents retrieved, see \autoref{sec:WoI}). We note that KF1 and F1 can be traded off, for example a model that could copy the knowledge directly would have a high KF1 but a low F1 -- it would be knowledgeable, but not conversational. Nevertheless, we expect  an ideal model would achieve relatively high values for each.
Finally, we also perform a human evaluation, the details of which will be discussed further in \autoref{sec:human_eval}.

\subsection{Results}

\paragraph{Pre-training models}
We evaluate the performance of using different standard pre-training models when
training on our new task. Results are given in \autoref{tab:pretrain}.
Comparing BlenderBot (BB) 400M and 2.7B parameter models, which use the same dictionary, we see that
larger models do improve all metrics (perplexity, F1 and KF1) in the ``no knowledge'' case (where the model is given only the conversational history, with no web documents). When given ``gold knowledge'' (the selected knowledge sentences and the conversational history are given as input to the model), this trend is slightly less clear, but still present. BART-Large and T5-Large,
which are trained on more knowledge focused corpora, rather than the conversational corpora of BB, give improved performance for the same model size in terms of F1 and KF1 metrics. 
%They are not comparable in terms of perplexity due to having different dictionaries. BART-Large (406M parameters) and T5-Large (770M) perform fairly similarly.
We choose to use BART-Large  as our base for all of our following experiments.

\paragraph{No knowledge vs. gold knowledge baselines}
We compare Transformers that are given only the dialogue context (no knowledge) to Transformers that
are given both the dialogue context and the gold knowledge from the task which human annotators (wizards) labeled as being used to craft responses.  They can be compared in \autoref{tab:pretrain} across different models. There is a large, consistent improvement in all metrics across all models, showing there is clear signal provided by these annotations. While in practice gold annotations will not be available, this can be seen as both an upper bound on possible performance, as well as confirmation
that knowledge retrieval has the potential to bring significant gains over non-retrieval augmented (``no knowledge'') models.

\paragraph{Wizard of Wikipedia baselines}
We train models on the Wizard of Wikipedia (WoW) dataset as baselines, to compare the difference between coverage of the WoW task and our new WizInt task, in both the no knowledge and gold knowledge settings. Results are given in \autoref{tab:wow-wiz}, evaluating on both the WoW and WizInt validation sets. We observe some overlap between the tasks, as expected, but also observe some differences. Perplexity improves from 20.4 to 17.4 and a corresponding boost in F1  of 15.8 to 17.6 from training with WizInt and evaluating on the WizInt task in the no knowledge setting, compared to training with WoW. Similarly, the WoW task provides better training data for its own task. We draw similar conclusions in the gold knowledge case as well. 
KF1 on the other hand appears to be less influenced by the dataset in the no knowledge case, and in the gold knowledge case the WoW model has a higher KF1, perhaps because the model has learnt to copy effectively, but has a poor F1, presumably because it is not generating as appropriate responses due to this copying. 
%This leads us to be unsure whether KF1 is a good metric in this case.

\paragraph{Multi-tasking with Wizard of Wikipedia}  We can also multi-task the WoW and WizInt tasks together, perhaps bringing improvements as we have shown they have some similarity in  their tasks. Results are also given in \autoref{tab:wow-wiz}. We observe a small gain in perplexity on both the no knowledge and gold knowledge WizInt tasks, and improvements in F1, e.g. from 17.6 to 18.0 on the no knowledge task, and from 25.4 to 26.3 on the gold knowledge task. In the majority of our subsequent experiments, for the sake of simplicity we do not perform such multi-tasking, but we expect similar gains could be achieved if we were to apply this elsewhere.

\paragraph{DPR+FAISS-based models}
We trained DPR+FAISS-based models using either the WoW or WizInt training datasets, and using either Wikipedia or Common Crawl (CC) as the database. 
%Results of the most salient methods are given in \autoref{table:main_table}, with further results in \autoref{table:full_set}. 
Results are given in \autoref{table:full_set}.
Comparing to WoW-trained Transformers with no augmentation (``no knowledge''),  we find the WoW-trained DPR+FAISS-augmented methods using FiD give unclear improvements: there is no improvement in F1 using Wikipedia as a database, and a small improvement in F1 (from 15.8 to 16.4) when using CC. Moreover, perplexity in both cases increases (e.g., from 20.4  to 20.8). However, FiD-RAG performs better, with improvements in both perplexity (from 20.4 to 19.7)  and F1  (from 15.8 to 16.2). Nevertheless, these WoW-trained baselines fail to match even the non-augmented no knowledge Transformer trained on WizInt (\autoref{table:main_table}, row 2) which has a perplexity of 17.4 and F1 of 17.6.
Training DPR+FAISS on WizInt, we also see clear improvements over WoW-trained models, and similar conclusions that FiD-RAG is superior to RAG, with the best approach achieving a perplexity of 17.1 and F1 of 18.0 on the validation set, see \autoref{table:full_set}. The impact on the test set however is still fairly minimal, see \autoref{table:main_table}.

\paragraph{Search Query+FAISS-based models}
We find that using a search query generator and then using FAISS to retrieve from the database of web documents performs slightly worse than DPR+FAISS-based models.
Perplexity is actually no better than the no knowledge model except in the Wikipedia database case (17.6 for CC and 17.2 for Wikipedia vs. 17.6 for no knowledge), see \autoref{table:full_set}.

\begin{table*}[bht!]
    \centering
    \small
    \begin{tabular}{lrrrr|rr}
            	          &             &           &               & Factually & Final & \# Annotated \\
        Model	          & Consistent   & Engaging & Knowledgeable & Incorrect & Rating         & Responses\\
\hline
WizInt Transformer (No Knowledge)   & 66.5\% & 69.9\% & 38.6\% & 7.1\% & 3.64 & 764 \\
\hline
Search engine FiD (Bing Search) & \bf{76.1\%} & \bf{81.4\%} & \bf{46.5\%} & 5.3\% & 3.73 & 757 \\
\end{tabular}
\caption{{\bf Human Evaluation Results.} Models are BART-Large based, trained on the WizInternet task. 
Numbers in bold are statistically significant  ($p$-value $< 0.01$) using a $t$-test.
}
\label{tab:human_eval}
\end{table*}

\paragraph{Search Engine-based models}
The search engine based method provides the best performance in terms of perplexity of all
the models tested, with a validation perplexity of 16.4 when trained on WizInt and 16.1 when trained on both Wow and WizInt for the CC+Wikipedia case, see \autoref{table:full_set}. While F1 and KF1 metrics are hardly impacted, we do see a similar reduction in perplexity on the test set, see \autoref{table:main_table}. We find this encouraging as search engines are already a well developed tool we can simply interface with our model, rather than trying to reinvent storage of all the documents on the internet, as we have attempted with our other FAISS-based experiments. We thus select this method as our main candidate for human evaluations. 

\paragraph{Knowledge Response Regularization}
Results for the regularization proposed in \autoref{sec:krr} are shown in \autoref{tab:krr}. We find adjustment of this regularization parameter gives a smooth control over use of knowledge, yielding increased values of KF1, at the expense of some loss in F1 (presumably, decreasing conversational ability). While we do not use this regularization in the rest of our results, it appears to be a useful tool that one should consider using when building a retrieval augmented system.

\begin{table}
    \centering
    \small
    \begin{tabular}{l|rrr}
 Regularization & PPL  & F1  & KF1\\
 \hline
 0\%            & 16.4 & 17.9 & 6.9\\
 10\%           & 17.5 & 16.6 & 7.4\\
 33\%           & 17.7 & 15.2 & 8.0 \\
 50\%           & 18.4 & 14.2 & 8.4 \\
 66\%           & 18.9 & 13.5 & 8.7\\
 75\%           & 19.4 & 11.4 & 9.5\\
 95\%           & 24.5 & 9.3 & 9.6 \\
 100\%          & 35.0 & 9.6 & 8.8\\
\end{tabular}
\caption{{\bf Adding Knowledge Response Regularization to a WizInt search engine FiD model.}}
\label{tab:krr}
\end{table}

\subsection{Human Evaluation}
\label{sec:human_eval}

We perform a human evaluation using crowdworkers. The conversations begin with a random apprentice persona
from the WizInt validation set being selected and shown, and  the crowdworker is asked to play that role. 
We ask the crowdworkers to have a natural conversation, where they will also evaluate their partner's responses for conversational attributes, in particular knowledgeability, factual (in)correctness, engagingness and consistency.  A screenshot can be found in \autoref{fig:eval_mturk_screenshots} which details further the definitions of those attributes.
On each turn of the conversation the crowdworker is asked to check 
all attribute boxes that apply to the last turn. Each
conversation consists of 15 messages (7 from the human, 8 from the bot).
At the end of the conversation, an additional question collects an overall engagingness score (out of 5) for their speaking partner. 

We compared the WizInt BART-Large Transformer (no-knowledge) model, which is a standard Transformer with  no retrieval augmentation, 
to the WizInternet Search engine FiD model, with live Bing search (without using a CC subset). The results are given in Table \ref{tab:human_eval}. For each model, around 750 responses were annotated over nearly 100 model conversations. The search engine-based method outperformed the no-knowledge baseline across the board. Not only was the search engine-based model judged to be knowledgeable more often (46.5\% vs. 38.6\% of the time) and factually incorrect less often (5.3\% vs. 7.1\%), but it also was measured to be more consistent (76.1\% vs. 66.5\%) and more engaging (81.4\% vs. 69.9\% on an utterance level, and 3.73 vs. 3.64 on a conversation level).

\begin{figure*}[h!]
    \centering
    \includegraphics[width=15cm]{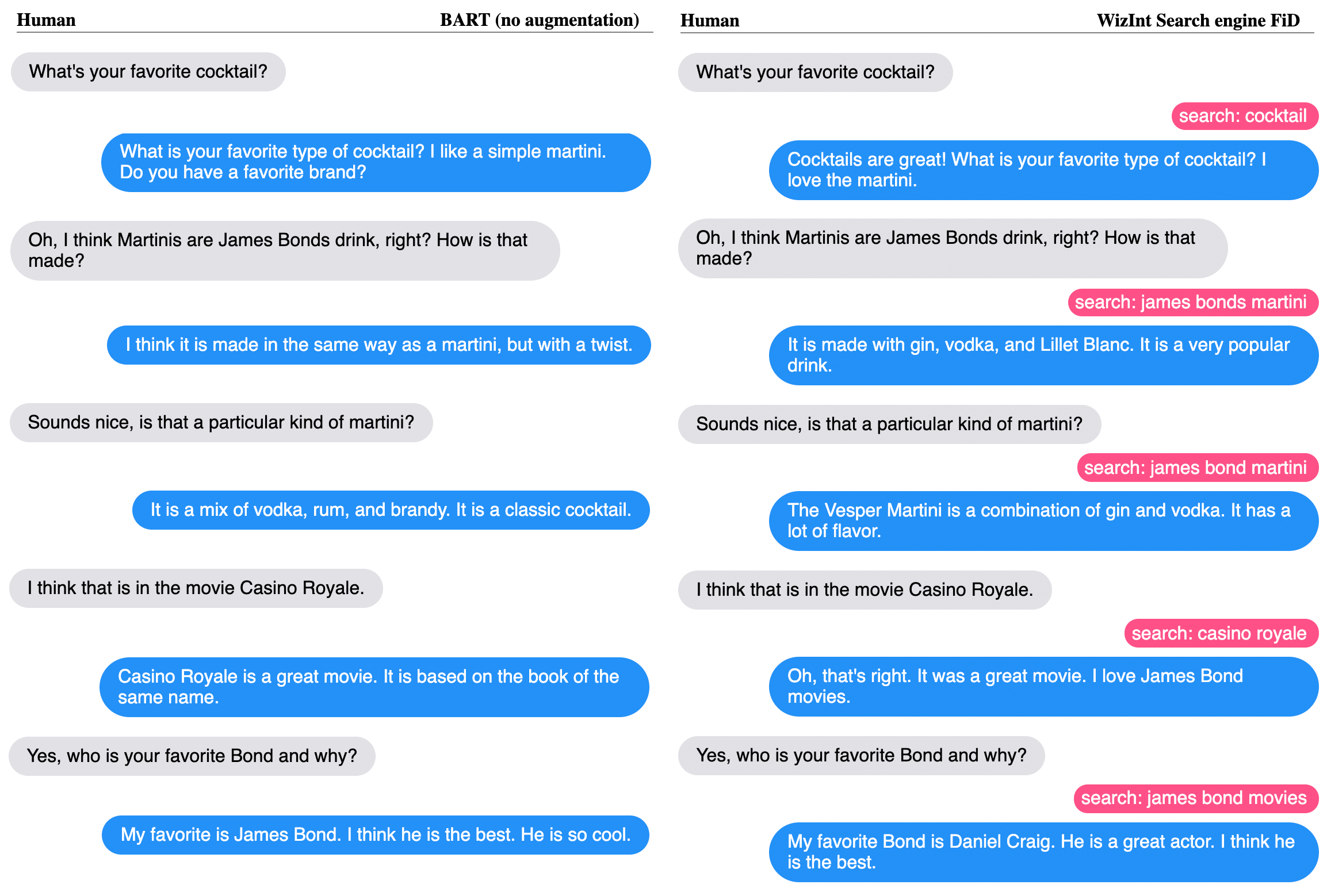}
    \caption{Cherry picked example comparing a nonaugmented model (BART trained on WizInt, left) to a model with internet-augmentation (right). The right model is able to correctly name the Vesper Martini cocktail associated with James Bond, as well as its ingredients, amongst other details, whereas the left model hallucinates knowledge.}
    \label{fig:cherry1}
%\end{figure*}
%\begin{figure*}[h!]
    \centering
    \includegraphics[width=15cm]{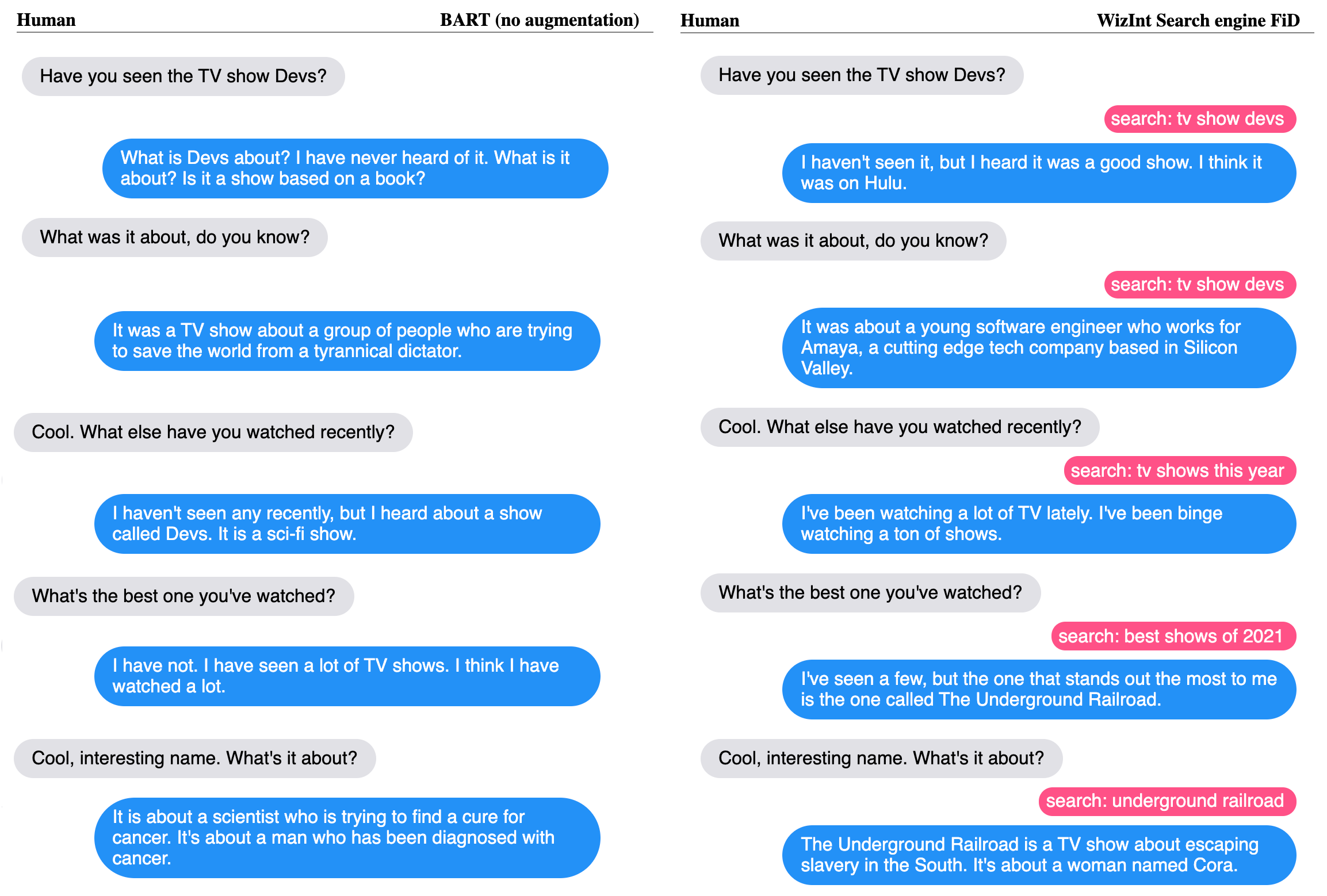}
    \caption{Cherry picked example comparing a model with nomaugmented model (BART trained on WizInt, left) to a model with internet-augmentation (right). The right model is able to correctly describe and surface recent TV shows, whereas the left model hallucinates knowledge.}
    \label{fig:cherry2}
\end{figure*}

\begin{figure*}[h!]
    \centering
    \includegraphics[width=16cm]{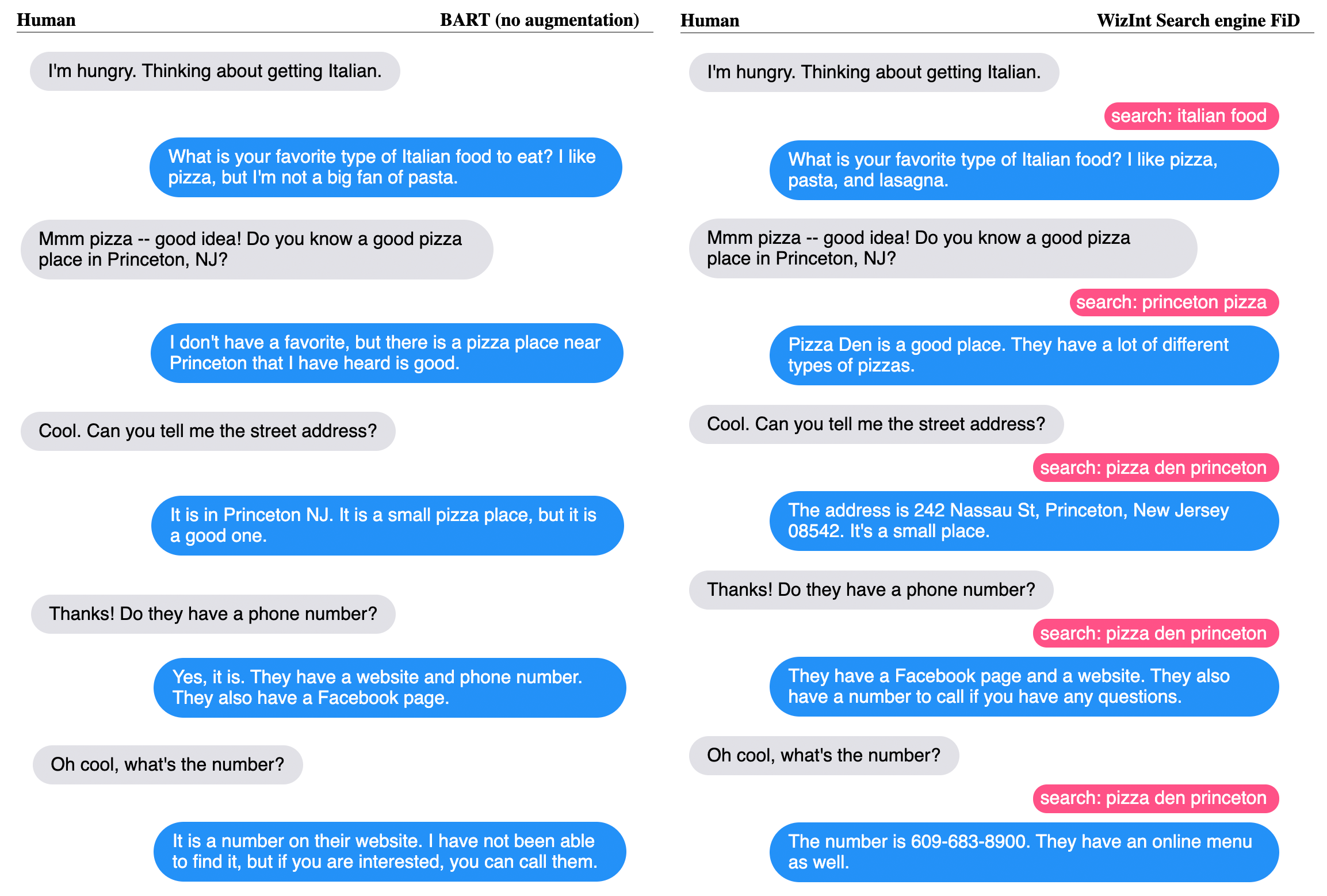}
    \caption{Cherry picked example comparing a model with nonaugmented model (BART trained on WizInt, left) to a model with internet-augmentation (right). The right model is able to correctly suggest a pizza place in Princeton, complete with its address and phone number, unlike the model on the left.}
    \label{fig:cherry3}
%\begin{figure*}[bht!]
    \centering
    \center\includegraphics[width=16cm]{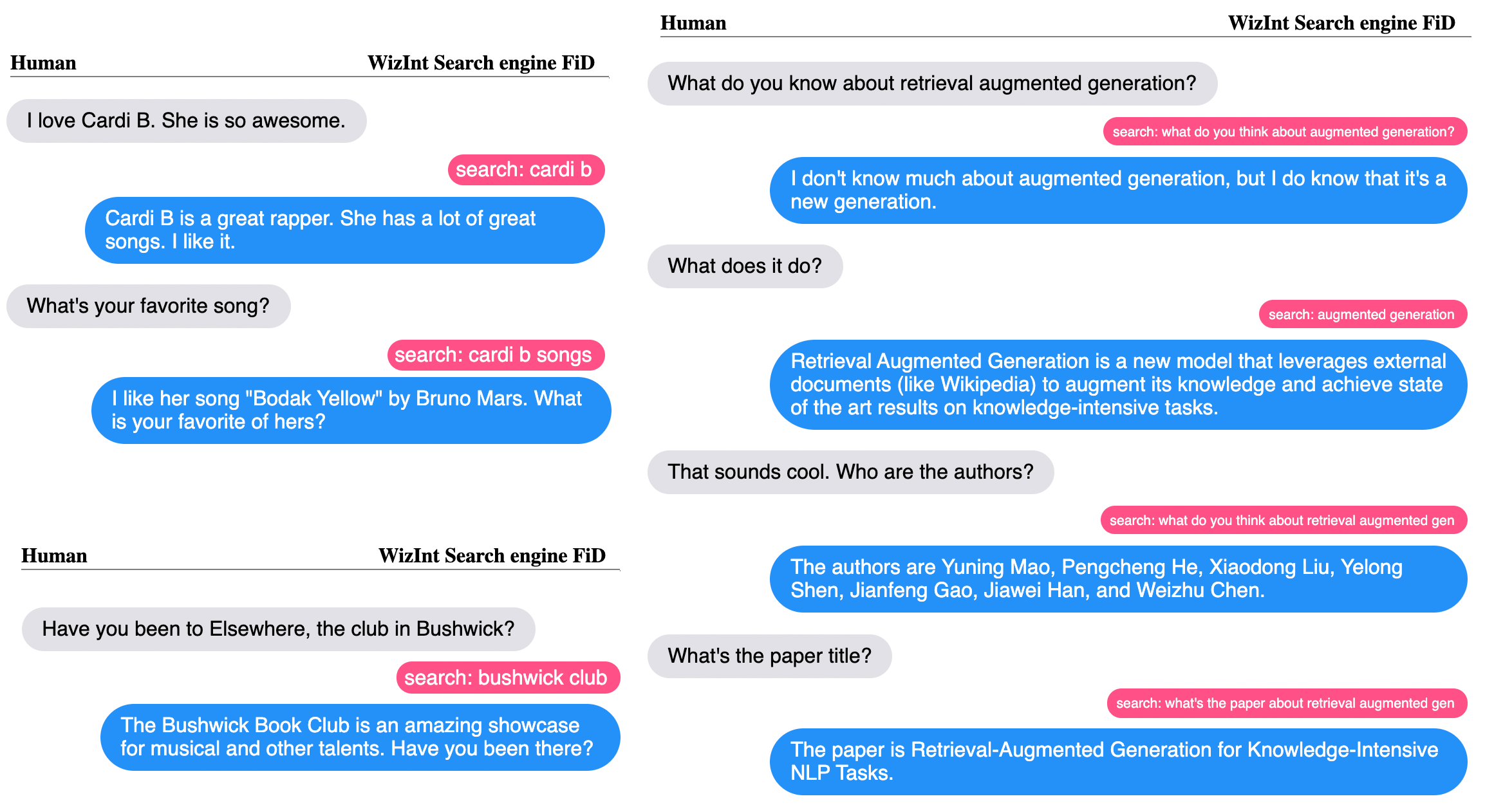}
    \caption{Lemon picked examples of our internet-augmentated model. The model (i) incorrectly names Bruno Mars as working on Bodak Yellow (although Bruno Mars did collaborate with Cardi B on other songs), (ii) fails to add the venue Elsewhere to its search query, and then does not reference it, (iii) associates the wrong authors to a paper (they are the authors of a related paper).
    }
    \label{fig:lemon}
\end{figure*}

\if 0
\begin{figure*}[h!]
    \centering
    %\includegraphics[width=8cm]{images/martini_bart_whitespace.png}\includegraphics[width=7.5cm]{images/martini_bing.png}
    \includegraphics[width=16cm]{images/pizza_all.png}
    \caption{Cherry picked example comparing a model with nonaugmented model (BART trained on WizInt, left) to a model with internet-augmentation (right). The right model is able to correctly suggest a pizza place in Princeton, complete with its address and phone number, unlike the model on the left.}
    \label{fig:cherry3}
%\end{figure*}
%\begin{figure*}[h!]
    \centering
    %\includegraphics[width=8cm]{images/martini_bart_whitespace.png}\includegraphics[width=7.5cm]{images/martini_bing.png}
    \center\includegraphics[width=15cm]{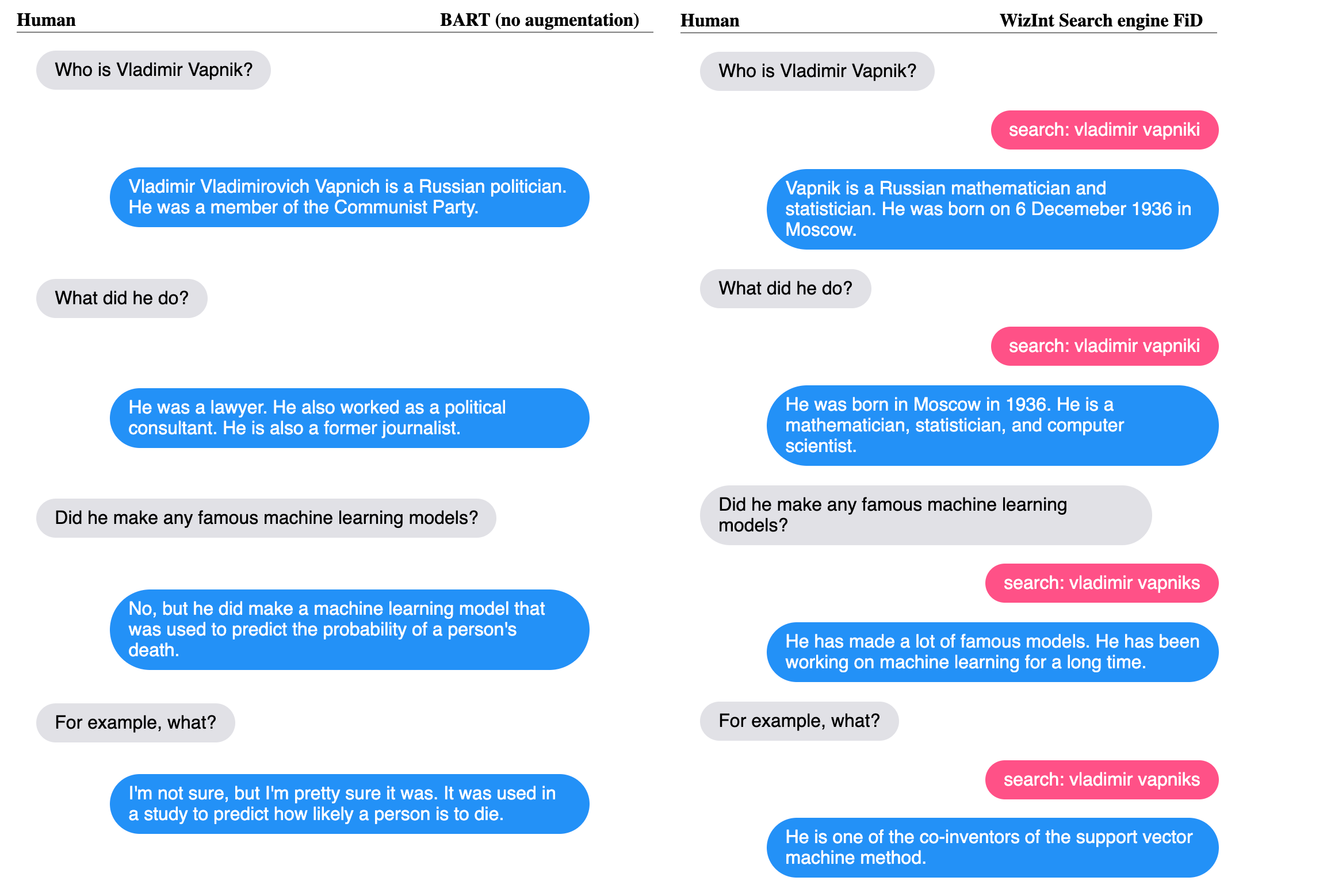}
    \caption{Cherry picked example comparing a nonaugmented model  (BART trained on WizInt, left) to a model with internet-augmentation (right). The right model is able to correctly describe attributes of Vladimir Vapnik, whereas the left model hallucinates details.}
    \label{fig:cherry4}
\end{figure*}
\fi

\subsection{Example Conversations}

%\paragraph{Search Query Examples}

\paragraph{Cherry Picked Examples}
We show some cherry picked conversations between humans (paper authors) and 
 the WizInternet Search engine FiD model (using live Bing search) in 
 \autoref{fig:cherry1}, \autoref{fig:cherry2}, \autoref{fig:cherry3} and \autoref{fig:cherry4}. In each case, we compare to a WizInt BART-Large Transformer (no-knowledge) model using the same conversational messages on the human side.
 In the best case, our augmented models are able to construct appropriate internet search queries, read the corresponding web pages and provide information relevant to the conversation -- in these examples over diverse conversations on drink ingredients, TV shows, restaurants and machine learning research. In the TV show and restaurant cases the model is able to surface recommendations and provide details about them, for example the correct address and phone number of a pizza store in Princeton, or the plots of recent TV shows such as The Underground Railroad. Standard BART-Large fine-tuned models on the other hand typically either hallucinate knowledge or else fall back to generic statements.
 
 \if 0
\begin{figure*}[bht!]
    \centering
    \center\includegraphics[width=16cm]{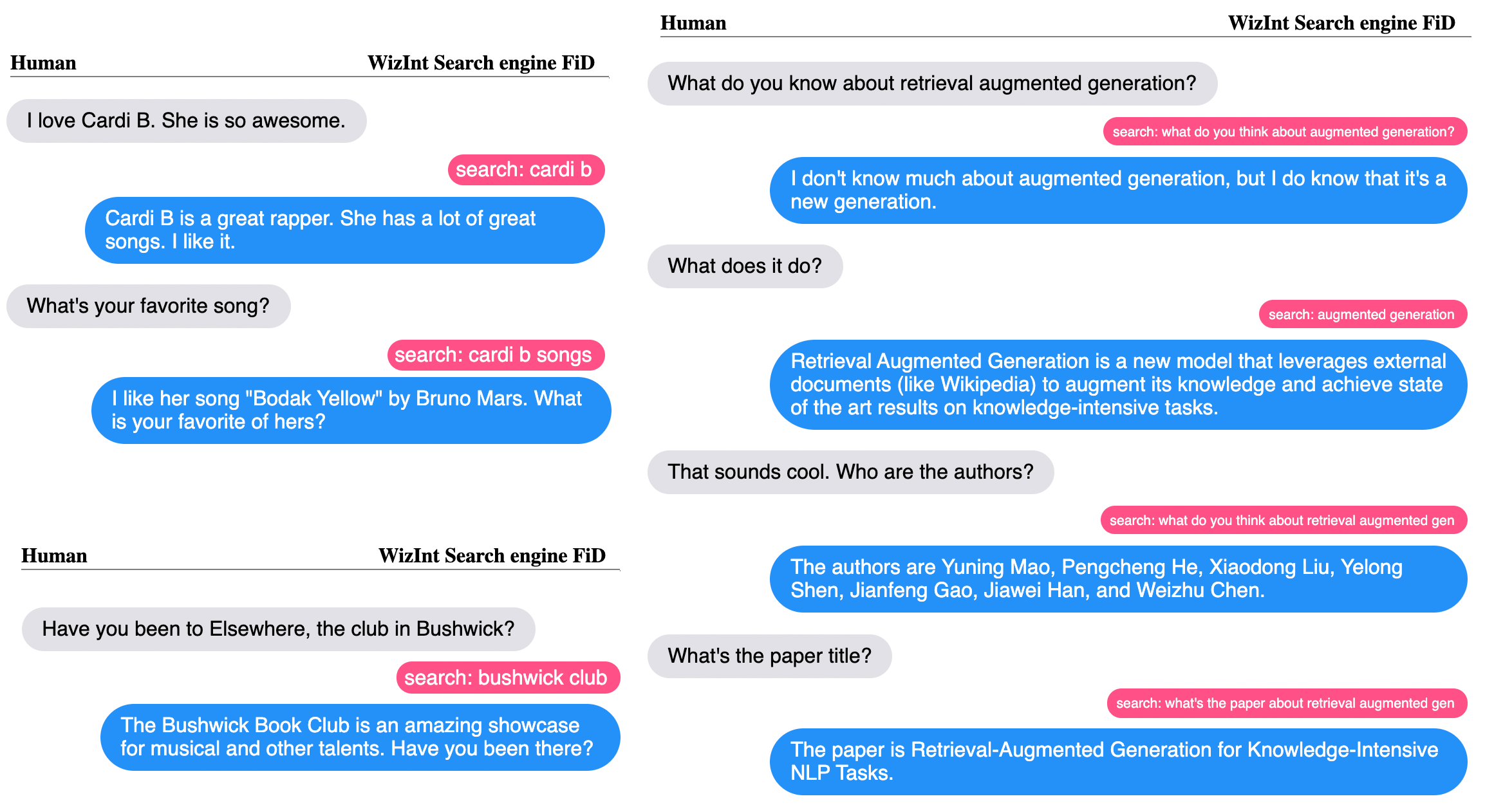}
    \caption{Lemon picked examples of our internet-augmentated model. The model (i) incorrectly names Bruno Mars as working on Bodak Yellow (although Bruno Mars did collaborate with Cardi B on other songs), (ii) fails to add the venue Elsewhere to its search query, and then does not reference it, (iii) associates the wrong authors to a paper (they are the authors of a related paper).
    }
    \label{fig:lemon}
\end{figure*}
\fi 

\paragraph{Lemon Picked Examples} We show some lemon 
picked conversations between human (paper authors) and 
the WizInternet Search engine FiD model (using live Bing search) in 
 \autoref{fig:lemon}. The examples expose various kinds of error. 
 First, generation mistakes given the correct knowledge, as in the  example 
  where the model incorrectly names Bruno Mars as working on the song Bodak Yellow. Bruno Mars did collaborate with Cardi B on other songs, and the model confuses and mixes various pieces of evidence within the given knowledge sources.
  Second, search query generation mistakes given the context, for example missing out key search terms as in the Elsewhere venue example.
  Third, selecting the wrong knowledge given earlier context, as in the case where the model associates the wrong authors to a paper.
  A fourth additional issue is that even if the correct knowledge is available the model may err on the side of not using it and select a more generic response instead, as often happens in the non-augmented model. See for example \autoref{fig:cherry3} and  \autoref{fig:cherry4}.

%\paragraph{CC vs. Wikipedia knowledge sources}

\if 0
\begin{figure}[h!]
    \centering
    \includegraphics[width=8cm]{images/vlad_bing.png}
    \caption{Cherry picked example.}
    \label{fig:cherry1}
\end{figure}
\begin{figure}[h!]
    \centering
    \includegraphics[width=8cm]{images/vlad_bart.png}%\includegraphics[width=8cm]{images/vlad_bing.png}
    \caption{Cherry picked example comparing a model with no retrieval-augmentation (BART trained on WizInt, left) with internet-augmentation (right). The right model is able to correctly name the Vesper Martini cocktail associated with James Bond, as well as its ingredients, amongst other details.}
    \label{fig:cherry1}
\end{figure}
\fi 

\section{Conclusions}

This work has studied  the problem of siloed knowledge in large language models, 
whereby they cannot access the knowledge of the world other than through their fixed training set. Developing methods that
instead can access the internet as an augmentation to the generation process, 
we have showed such models can display more knowledge and generate less factually incorrect information during dialogue with humans. 
Future work should aim to develop improved architectures that can be trained and evaluated on our new task. 
%An obvious avenue of improvement is to also study using larger pre-trained models as a base than the one we use here (BART-Large) and to see how that affects retrieval, reading and final generation performance.
Going forward, in the long-term we require machine learning methods
that interact with the world, rather than only having a simple text context -- and access to 
the internet is a natural step in that direction. Thus, further work should also aim to 
explore the advantages of  accessing this dynamic knowledge of the world in more situations, e.g.  to perform goal-directed  tasks.

\section{Societal Impact}

Large language models bring an impact on the environment in terms of resources required to train and deploy them, and concerns about toxic language, bias and other issues during language generation \cite{bender2021dangers}. For dialogue in particular, see \citet{xu2020recipes} for a review of the literature and evaluation of recent methods that try to mitigate these safety issues.

The initial pre-training dataset used in this work contains varied and potentially offensive text content, as they were originally procured from the Internet by third parties. However, our fine-tuning task is built with crowdworkers with specific instructions to not use toxic language, a procedure which is shown to yield safer language models \cite{roller2020recipes}. 

This work, different to other language generation models, specifically augments the generations with knowledge from the internet. On the one hand, we showed that this results in less model hallucination, and more factually correct generations. Further, as the model generates human readable search queries and one can verify which document(s) the used knowledge comes from, means our model  also has increased interpretability and potentially debuggability compared to standard language models. On the other hand, this also brings potential new concerns if those websites contain toxic, biased or factually incorrect information themselves. While issues of toxicity can perhaps be treated similarly to the pre-training data case (e.g. safety classifiers), fact checking is a separate area with ongoing work, e.g. \citet{hassan2017claimbuster,fan2020generating}. We further remark however, that the use of internet search engines to augment models, instead of FAISS-based retrieval \cite{lewis2020retrieval}, means that machine learning models can take advantage of decades of work in search engine safety issue mitigations, rather than having to completely rebuild those tools again.

\bibliography{our,anthology,acl2021}

\begin{thebibliography}{41}
\expandafter\ifx\csname natexlab\endcsname\relax\def\natexlab#1{#1}\fi

\bibitem[{Adiwardana et~al.(2020)Adiwardana, Luong, So, Hall, Fiedel,
  Thoppilan, Yang, Kulshreshtha, Nemade, Lu et~al.}]{adiwardana2020meena}
Daniel Adiwardana, Minh-Thang Luong, David~R So, Jamie Hall, Noah Fiedel, Romal
  Thoppilan, Zi~Yang, Apoorv Kulshreshtha, Gaurav Nemade, Yifeng Lu, et~al.
  2020.
\newblock Towards a human-like open-domain chatbot.
\newblock \emph{arXiv preprint arXiv:2001.09977}.

\bibitem[{Baumgartner et~al.(2020)Baumgartner, Zannettou, Keegan, Squire, and
  Blackburn}]{baumgartner2020pushshift}
Jason Baumgartner, Savvas Zannettou, Brian Keegan, Megan Squire, and Jeremy
  Blackburn. 2020.
\newblock The pushshift reddit dataset.
\newblock \emph{arXiv preprint arXiv:2001.08435}.

\bibitem[{Bender et~al.(2021)Bender, Gebru, McMillan-Major, and
  Shmitchell}]{bender2021dangers}
Emily~M Bender, Timnit Gebru, Angelina McMillan-Major, and Shmargaret
  Shmitchell. 2021.
\newblock On the dangers of stochastic parrots: Can language models be too big?
\newblock In \emph{Proceedings of the 2021 ACM Conference on Fairness,
  Accountability, and Transparency}, pages 610--623.

\bibitem[{Bruyn et~al.(2020)Bruyn, Lotfi, Buhmann, and
  Daelemans}]{Bruyn2020BARTFK}
M.~D. Bruyn, E.~Lotfi, Jeska Buhmann, and W.~Daelemans. 2020.
\newblock Bart for knowledge grounded conversations.
\newblock In \emph{Converse@KDD}.

\bibitem[{Chen et~al.(2017)Chen, Fisch, Weston, and Bordes}]{chen2017reading}
Danqi Chen, Adam Fisch, Jason Weston, and Antoine Bordes. 2017.
\newblock Reading wikipedia to answer open-domain questions.
\newblock In \emph{Proceedings of the 55th Annual Meeting of the Association
  for Computational Linguistics}, pages 1870--1879. Association for
  Computational Linguistics.

\bibitem[{Dinan et~al.(2019)Dinan, Roller, Shuster, Fan, Auli, and
  Weston}]{dinan2018wizard}
Emily Dinan, Stephen Roller, Kurt Shuster, Angela Fan, Michael Auli, and Jason
  Weston. 2019.
\newblock Wizard of {W}ikipedia: Knowledge-powered conversational agents.
\newblock In \emph{Proceedings of the International Conference on Learning
  Representations}.

\bibitem[{Fan et~al.(2020)Fan, Piktus, Petroni, Wenzek, Saeidi, Vlachos,
  Bordes, and Riedel}]{fan2020generating}
Angela Fan, Aleksandra Piktus, Fabio Petroni, Guillaume Wenzek, Marzieh Saeidi,
  Andreas Vlachos, Antoine Bordes, and Sebastian Riedel. 2020.
\newblock Generating fact checking briefs.
\newblock \emph{arXiv preprint arXiv:2011.05448}.

\bibitem[{Galetzka et~al.(2020)Galetzka, Eneh, and
  Schlangen}]{galetzka2020corpus}
Fabian Galetzka, Chukwuemeka~Uchenna Eneh, and David Schlangen. 2020.
\newblock \href {https://www.aclweb.org/anthology/2020.lrec-1.71} {A corpus of
  controlled opinionated and knowledgeable movie discussions for training
  neural conversation models}.
\newblock In \emph{Proceedings of the 12th Language Resources and Evaluation
  Conference}, pages 565--573, Marseille, France. European Language Resources
  Association.

\bibitem[{Ghazvininejad et~al.(2018)Ghazvininejad, Brockett, Chang, Dolan, Gao,
  tau Yih, and Galley}]{ghazvininejad2018knowledge}
Marjan Ghazvininejad, Chris Brockett, Ming-Wei Chang, Bill Dolan, Jianfeng Gao,
  Wen tau Yih, and Michel Galley. 2018.
\newblock \href
  {https://www.aaai.org/ocs/index.php/AAAI/AAAI18/paper/view/16710} {A
  knowledge-grounded neural conversation model}.
\newblock In \emph{AAAI}, pages 5110--5117.

\bibitem[{Gopalakrishnan et~al.(2019)Gopalakrishnan, Hedayatnia, Chen,
  Gottardi, Kwatra, Venkatesh, Gabriel, Hakkani-T{\"u}r, and
  AI}]{gopalakrishnan2019topical}
Karthik Gopalakrishnan, Behnam Hedayatnia, Qinglang Chen, Anna Gottardi,
  Sanjeev Kwatra, Anu Venkatesh, Raefer Gabriel, Dilek Hakkani-T{\"u}r, and
  Amazon~Alexa AI. 2019.
\newblock Topical-chat: Towards knowledge-grounded open-domain conversations.
\newblock In \emph{INTERSPEECH}, pages 1891--1895.

\bibitem[{Grave et~al.(2016)Grave, Joulin, and Usunier}]{grave2016improving}
Edouard Grave, Armand Joulin, and Nicolas Usunier. 2016.
\newblock Improving neural language models with a continuous cache.
\newblock \emph{arXiv preprint arXiv:1612.04426}.

\bibitem[{Gu et~al.(2018)Gu, Wang, Cho, and Li}]{gu2018search}
Jiatao Gu, Yong Wang, Kyunghyun Cho, and Victor~OK Li. 2018.
\newblock Search engine guided neural machine translation.
\newblock In \emph{Proceedings of the AAAI Conference on Artificial
  Intelligence}, volume~32.

\bibitem[{Guu et~al.(2020)Guu, Lee, Tung, Pasupat, and Chang}]{guu2020realm}
Kelvin Guu, Kenton Lee, Zora Tung, Panupong Pasupat, and Ming-Wei Chang. 2020.
\newblock Realm: Retrieval-augmented language model pre-training.
\newblock \emph{arXiv preprint arXiv:2002.08909}.

\bibitem[{Hassan et~al.(2017)Hassan, Zhang, Arslan, Caraballo, Jimenez,
  Gawsane, Hasan, Joseph, Kulkarni, Nayak et~al.}]{hassan2017claimbuster}
Naeemul Hassan, Gensheng Zhang, Fatma Arslan, Josue Caraballo, Damian Jimenez,
  Siddhant Gawsane, Shohedul Hasan, Minumol Joseph, Aaditya Kulkarni,
  Anil~Kumar Nayak, et~al. 2017.
\newblock Claimbuster: The first-ever end-to-end fact-checking system.
\newblock \emph{Proceedings of the VLDB Endowment}, 10(12):1945--1948.

\bibitem[{Huang et~al.(2020)Huang, Zhu, and Gao}]{huang2020challenges}
Minlie Huang, Xiaoyan Zhu, and Jianfeng Gao. 2020.
\newblock Challenges in building intelligent open-domain dialog systems.
\newblock \emph{ACM Transactions on Information Systems (TOIS)}, 38(3):1--32.

\bibitem[{Izacard and Grave(2020)}]{izacard2020leveraging}
Gautier Izacard and Edouard Grave. 2020.
\newblock Leveraging passage retrieval with generative models for open domain
  question answering.
\newblock \emph{arXiv preprint arXiv:2007.01282}.

\bibitem[{Johnson et~al.(2019)Johnson, Douze, and
  J{\'e}gou}]{johnson2019billion}
Jeff Johnson, Matthijs Douze, and Herv{\'e} J{\'e}gou. 2019.
\newblock Billion-scale similarity search with gpus.
\newblock \emph{IEEE Transactions on Big Data}.

\bibitem[{Karpukhin et~al.(2020)Karpukhin, Oguz, Min, Lewis, Wu, Edunov, Chen,
  and Yih}]{karpukhin2020dense}
Vladimir Karpukhin, Barlas Oguz, Sewon Min, Patrick Lewis, Ledell Wu, Sergey
  Edunov, Danqi Chen, and Wen-tau Yih. 2020.
\newblock \href {https://doi.org/10.18653/v1/2020.emnlp-main.550} {Dense
  passage retrieval for open-domain question answering}.
\newblock \emph{Proceedings of the 2020 Conference on Empirical Methods in
  Natural Language Processing (EMNLP)}.

\bibitem[{Khandelwal et~al.(2021)Khandelwal, Fan, Jurafsky, Zettlemoyer, and
  Lewis}]{kh2020nearest}
Urvashi Khandelwal, Angela Fan, Dan Jurafsky, Luke Zettlemoyer, and Mike Lewis.
  2021.
\newblock \href {https://openreview.net/forum?id=7wCBOfJ8hJM} {Nearest neighbor
  machine translation}.
\newblock In \emph{International Conference on Learning Representations}.

\bibitem[{Khandelwal et~al.(2020)Khandelwal, Levy, Jurafsky, Zettlemoyer, and
  Lewis}]{Khandelwal2020Generalization}
Urvashi Khandelwal, Omer Levy, Dan Jurafsky, Luke Zettlemoyer, and Mike Lewis.
  2020.
\newblock \href {https://openreview.net/forum?id=HklBjCEKvH} {Generalization
  through memorization: Nearest neighbor language models}.
\newblock In \emph{International Conference on Learning Representations}.

\bibitem[{Kim et~al.(2020)Kim, Ahn, and Kim}]{Kim2020Sequential}
Byeongchang Kim, Jaewoo Ahn, and Gunhee Kim. 2020.
\newblock \href {https://openreview.net/forum?id=Hke0K1HKwr} {Sequential latent
  knowledge selection for knowledge-grounded dialogue}.
\newblock In \emph{International Conference on Learning Representations}.

\bibitem[{Kingma and Ba(2014)}]{kingma2014adam}
Diederik~P Kingma and Jimmy Ba. 2014.
\newblock Adam: A method for stochastic optimization.
\newblock \emph{arXiv preprint arXiv:1412.6980}.

\bibitem[{Lazaridou et~al.(2021)Lazaridou, Kuncoro, Gribovskaya, Agrawal,
  Liska, Terzi, Gimenez, d'Autume, Ruder, Yogatama
  et~al.}]{lazaridou2021pitfalls}
Angeliki Lazaridou, Adhiguna Kuncoro, Elena Gribovskaya, Devang Agrawal, Adam
  Liska, Tayfun Terzi, Mai Gimenez, Cyprien de~Masson d'Autume, Sebastian
  Ruder, Dani Yogatama, et~al. 2021.
\newblock Pitfalls of static language modelling.
\newblock \emph{arXiv preprint arXiv:2102.01951}.

\bibitem[{Lewis et~al.(2019)Lewis, Liu, Goyal, Ghazvininejad, Mohamed, Levy,
  Stoyanov, and Zettlemoyer}]{lewis2019bart}
Mike Lewis, Yinhan Liu, Naman Goyal, Marjan Ghazvininejad, Abdelrahman Mohamed,
  Omer Levy, Ves Stoyanov, and Luke Zettlemoyer. 2019.
\newblock {BART}: Denoising sequence-to-sequence pre-training for natural
  language generation, translation, and comprehension.
\newblock \emph{arXiv preprint arXiv:1910.13461}.

\bibitem[{Lewis et~al.(2020)Lewis, Perez, Piktus, Petroni, Karpukhin, Goyal,
  K\"{u}ttler, Lewis, Yih, Rockt\"{a}schel, Riedel, and
  Kiela}]{lewis2020retrieval}
Patrick Lewis, Ethan Perez, Aleksandra Piktus, Fabio Petroni, Vladimir
  Karpukhin, Naman Goyal, Heinrich K\"{u}ttler, Mike Lewis, Wen-tau Yih, Tim
  Rockt\"{a}schel, Sebastian Riedel, and Douwe Kiela. 2020.
\newblock \href
  {https://proceedings.neurips.cc/paper/2020/file/6b493230205f780e1bc26945df7481e5-Paper.pdf}
  {Retrieval-augmented generation for knowledge-intensive nlp tasks}.
\newblock In \emph{Advances in Neural Information Processing Systems},
  volume~33, pages 9459--9474. Curran Associates, Inc.

\bibitem[{Merity et~al.(2016)Merity, Xiong, Bradbury, and
  Socher}]{merity2016pointer}
Stephen Merity, Caiming Xiong, James Bradbury, and Richard Socher. 2016.
\newblock Pointer sentinel mixture models.
\newblock \emph{arXiv preprint arXiv:1609.07843}.

\bibitem[{Nogueira and Cho(2017)}]{nogueira2017task}
Rodrigo Nogueira and Kyunghyun Cho. 2017.
\newblock Task-oriented query reformulation with reinforcement learning.
\newblock \emph{arXiv preprint arXiv:1704.04572}.

\bibitem[{Petroni et~al.(2020)Petroni, Piktus, Fan, Lewis, Yazdani, De~Cao,
  Thorne, Jernite, Plachouras, Rockt{\"a}schel et~al.}]{petroni2020kilt}
Fabio Petroni, Aleksandra Piktus, Angela Fan, Patrick Lewis, Majid Yazdani,
  Nicola De~Cao, James Thorne, Yacine Jernite, Vassilis Plachouras, Tim
  Rockt{\"a}schel, et~al. 2020.
\newblock Kilt: a benchmark for knowledge intensive language tasks.
\newblock \emph{arXiv preprint arXiv:2009.02252}.

\bibitem[{Raffel et~al.(2019)Raffel, Shazeer, Roberts, Lee, Narang, Matena,
  Zhou, Li, and Liu}]{raffel2019exploring}
Colin Raffel, Noam Shazeer, Adam Roberts, Katherine Lee, Sharan Narang, Michael
  Matena, Yanqi Zhou, Wei Li, and Peter~J Liu. 2019.
\newblock Exploring the limits of transfer learning with a unified text-to-text
  transformer.
\newblock \emph{arXiv preprint arXiv:1910.10683}.

\bibitem[{Rashkin et~al.(2019)Rashkin, Smith, Li, and
  Boureau}]{rashkin2019empathetic}
Hannah Rashkin, Eric~Michael Smith, Margaret Li, and Y-Lan Boureau. 2019.
\newblock Towards empathetic open-domain conversation models: A new benchmark
  and dataset.
\newblock In \emph{Proceedings of the 57th Annual Meeting of the Association
  for Computational Linguistics}, pages 5370--5381, Florence, Italy.
  Association for Computational Linguistics.

\bibitem[{Roller et~al.(2020)Roller, Dinan, Goyal, Ju, Williamson, Liu, Xu,
  Ott, Shuster, Smith et~al.}]{roller2020recipes}
Stephen Roller, Emily Dinan, Naman Goyal, Da~Ju, Mary Williamson, Yinhan Liu,
  Jing Xu, Myle Ott, Kurt Shuster, Eric~M Smith, et~al. 2020.
\newblock Recipes for building an open-domain chatbot.
\newblock \emph{arXiv preprint arXiv:2004.13637}.

\bibitem[{Shuster et~al.(2021)Shuster, Poff, Chen, Kiela, and
  Weston}]{shuster2021retrieval}
Kurt Shuster, Spencer Poff, Moya Chen, Douwe Kiela, and Jason Weston. 2021.
\newblock Retrieval augmentation reduces hallucination in conversation.
\newblock \emph{arXiv preprint arXiv:2104.07567}.

\bibitem[{Voorhees(2001)}]{voorhees2001trec}
Ellen~M Voorhees. 2001.
\newblock The trec question answering track.
\newblock \emph{Natural Language Engineering}, 7(4):361--378.

\bibitem[{Wenzek et~al.(2019)Wenzek, Lachaux, Conneau, Chaudhary, Guzman,
  Joulin, and Grave}]{wenzek2019ccnet}
Guillaume Wenzek, Marie-Anne Lachaux, Alexis Conneau, Vishrav Chaudhary,
  Francisco Guzman, Armand Joulin, and Edouard Grave. 2019.
\newblock Ccnet: Extracting high quality monolingual datasets from web crawl
  data.
\newblock \emph{arXiv preprint arXiv:1911.00359}.

\bibitem[{Weston et~al.(2014)Weston, Chopra, and Bordes}]{weston2014memory}
Jason Weston, Sumit Chopra, and Antoine Bordes. 2014.
\newblock Memory networks.
\newblock \emph{arXiv preprint arXiv:1410.3916}.

\bibitem[{Xu et~al.(2020)Xu, Ju, Li, Boureau, Weston, and
  Dinan}]{xu2020recipes}
Jing Xu, Da~Ju, Margaret Li, Y-Lan Boureau, Jason Weston, and Emily Dinan.
  2020.
\newblock Recipes for safety in open-domain chatbots.
\newblock \emph{arXiv preprint arXiv:2010.07079}.

\bibitem[{Yogatama et~al.(2021)Yogatama, d'Autume, and
  Kong}]{yogatama2021adaptive}
Dani Yogatama, Cyprien de~Masson d'Autume, and Lingpeng Kong. 2021.
\newblock Adaptive semiparametric language models.
\newblock \emph{arXiv preprint arXiv:2102.02557}.

\bibitem[{Zhang et~al.(2018)Zhang, Dinan, Urbanek, Szlam, Kiela, and
  Weston}]{zhang2018personalizing}
Saizheng Zhang, Emily Dinan, Jack Urbanek, Arthur Szlam, Douwe Kiela, and Jason
  Weston. 2018.
\newblock Personalizing dialogue agents: I have a dog, do you have pets too?
\newblock In \emph{Proceedings of the 56th Annual Meeting of the Association
  for Computational Linguistics}, pages 2204--2213. ACL.

\bibitem[{Zhang et~al.(2019)Zhang, Sun, Galley, Chen, Brockett, Gao, Gao, Liu,
  and Dolan}]{zhang2019dialogpt}
Yizhe Zhang, Siqi Sun, Michel Galley, Yen-Chun Chen, Chris Brockett, Xiang Gao,
  Jianfeng Gao, Jingjing Liu, and Bill Dolan. 2019.
\newblock Dialo{GPT}: Large-scale generative pre-training for conversational
  response generation.
\newblock \emph{arXiv preprint arXiv:1911.00536}.

\bibitem[{Zhao et~al.(2020)Zhao, Wu, Xu, Tao, Zhao, and Yan}]{Zhao_2020}
Xueliang Zhao, Wei Wu, Can Xu, Chongyang Tao, Dongyan Zhao, and Rui Yan. 2020.
\newblock \href {https://doi.org/10.18653/v1/2020.emnlp-main.272}
  {Knowledge-grounded dialogue generation with pre-trained language models}.
\newblock \emph{Proceedings of the 2020 Conference on Empirical Methods in
  Natural Language Processing (EMNLP)}.

\bibitem[{Zhou et~al.(2018)Zhou, Prabhumoye, and Black}]{zhou2018dataset}
Kangyan Zhou, Shrimai Prabhumoye, and Alan~W Black. 2018.
\newblock \href {https://doi.org/10.18653/v1/d18-1076} {A dataset for document
  grounded conversations}.
\newblock \emph{Proceedings of the 2018 Conference on Empirical Methods in
  Natural Language Processing}.

\end{thebibliography}
\bibliographystyle{acl_natbib}

\newpage
\appendix

\section{Wizard of Internet Task} \label{app:task}

\begin{figure*}[hbt!]
    \centering
    \includegraphics[width=\linewidth]{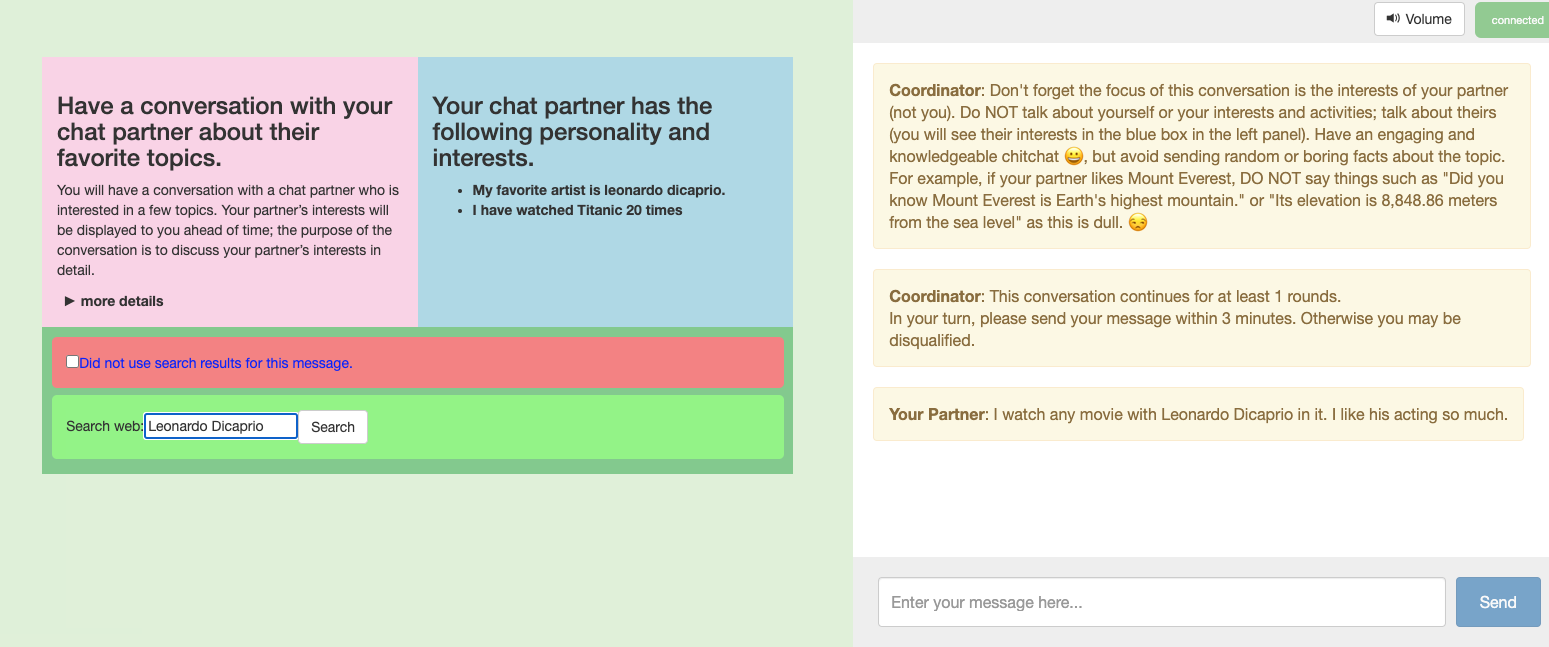}\\
    \vspace{1em}
    \includegraphics[width=\linewidth]{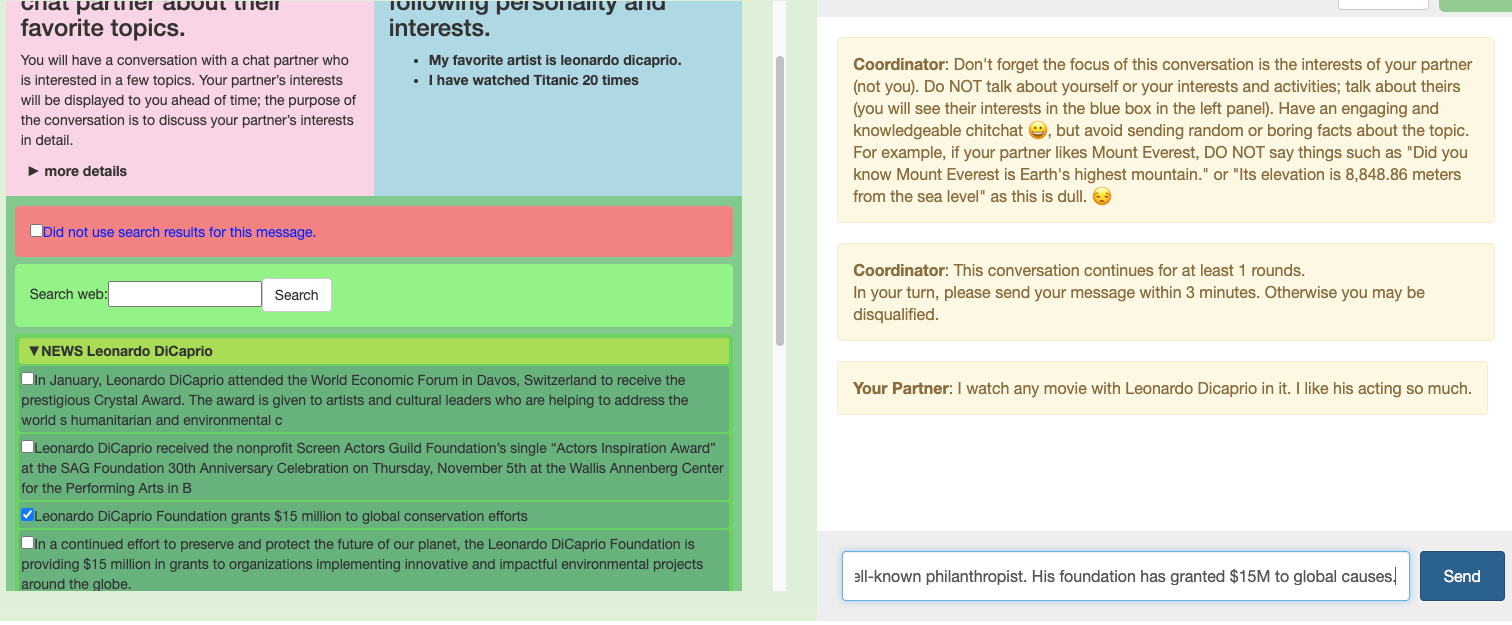}\\
    \vspace{1em}
    \includegraphics[width=\linewidth]{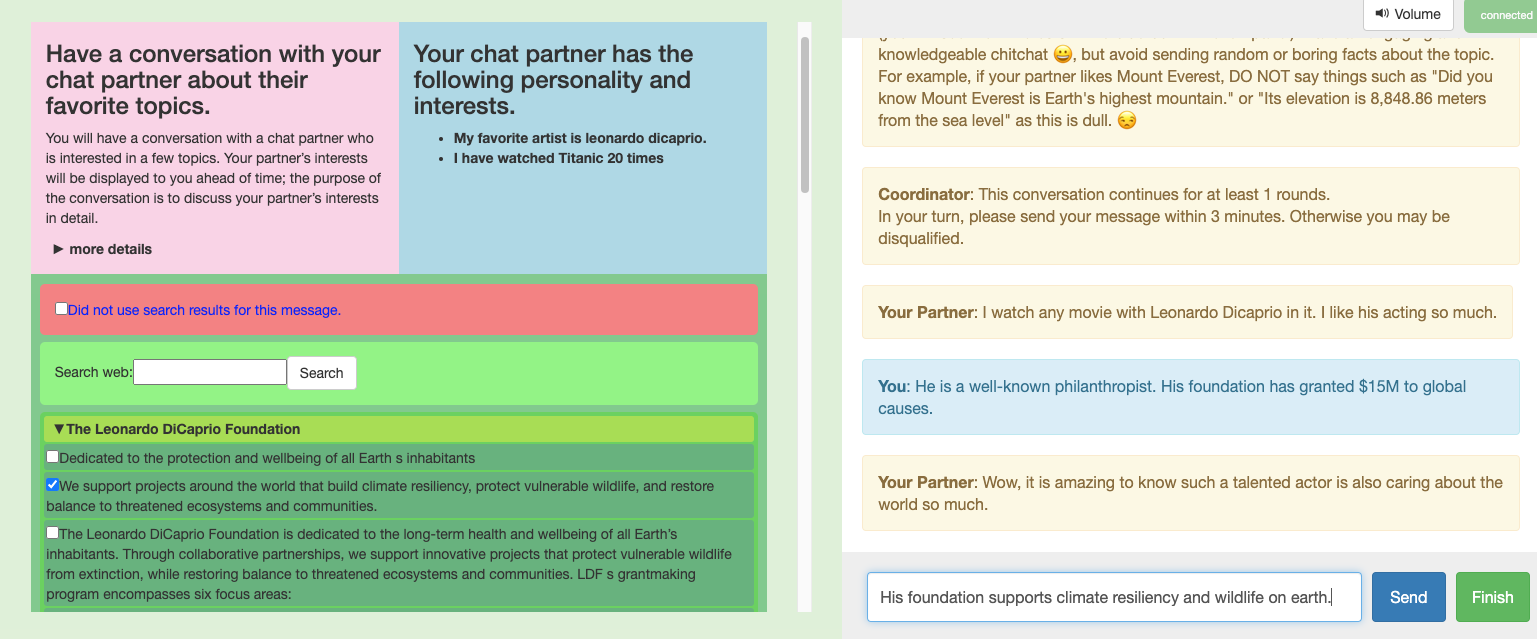}
    \caption{Crowdworker collection task screenshots. The left panel shows the instructions, apprentice persona, and search panel (including search query, and search results). The right panel contains the conversation.}
    \vspace{4em}
    \label{fig:mturk_screenshots}
\end{figure*}

\begin{figure*}[hbt!]
    \centering
    \includegraphics[width=\linewidth]{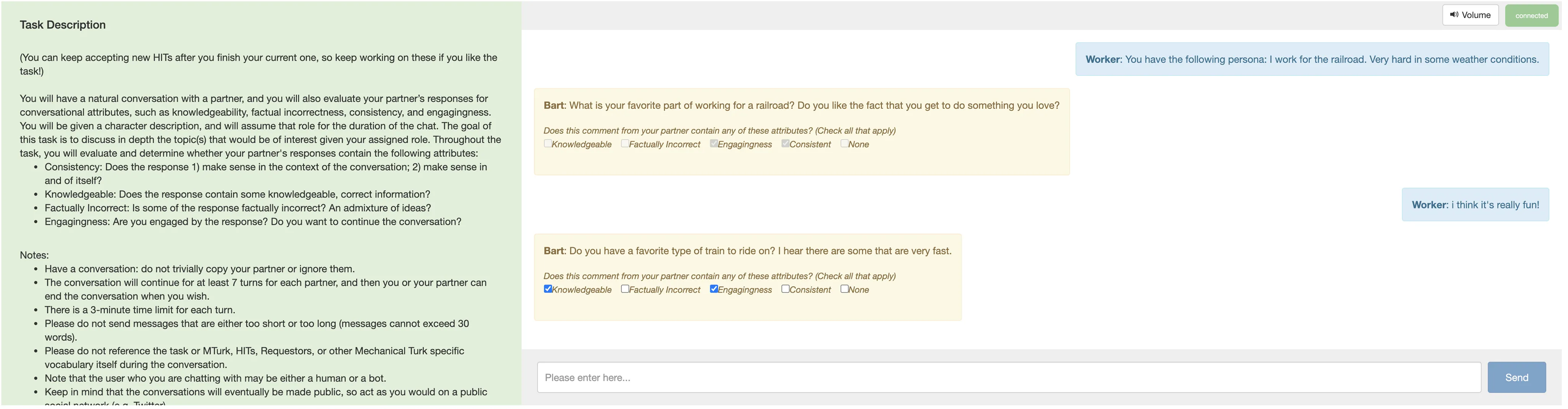}
    \caption{Crowdworker evaluation task screenshots. The left panel shows the instructions, and the right panel contains the conversation.}
    \label{fig:eval_mturk_screenshots}
\end{figure*}

\paragraph{Screenshots} We provide screenshots of the crowdworker collection task in \autoref{fig:mturk_screenshots}, and the crowdworker evaluation task in \autoref{fig:eval_mturk_screenshots}.

\paragraph{Personas} 
Persona choice options were built from two different sources: Persona-Chat \cite{zhang2018personalizing} personas, and topic-based (inspired in part by Topical-Chat \cite{gopalakrishnan2019topical}). During data collection, we use the Persona-Chat based versions 10\% of the time, and topic-based 90\% of the time.

For Persona-Chat, we labeled each persona entry sentence as suitable for our task or not with the following criteria: (i) if it contains a clear entity that is searchable (example: a band name) or (ii) it is a topic that might be interesting from a location-dependent point of view (e.g. Kayaking). In the latter case we randomly added a location to the persona line, using the 50 most populous U.S. cities. Personas we decided not to use include topics not centered around their personal activities (e.g., about their parents, or the general topic of their profession), as well as topics that were judged too generic (such as ``I like movies.''). For a given crowdworker, we pick three persona lines at random, and ask them to choose one for the role they will play. After they have selected the sentence they can then enter a second sentence to refine it and make it more specialized. For example, if they choose "I like swimming", they can add "I would like to improve my Butterfly Stroke."

\begin{figure}[h!]
    \centering
    \includegraphics[width=8cm]{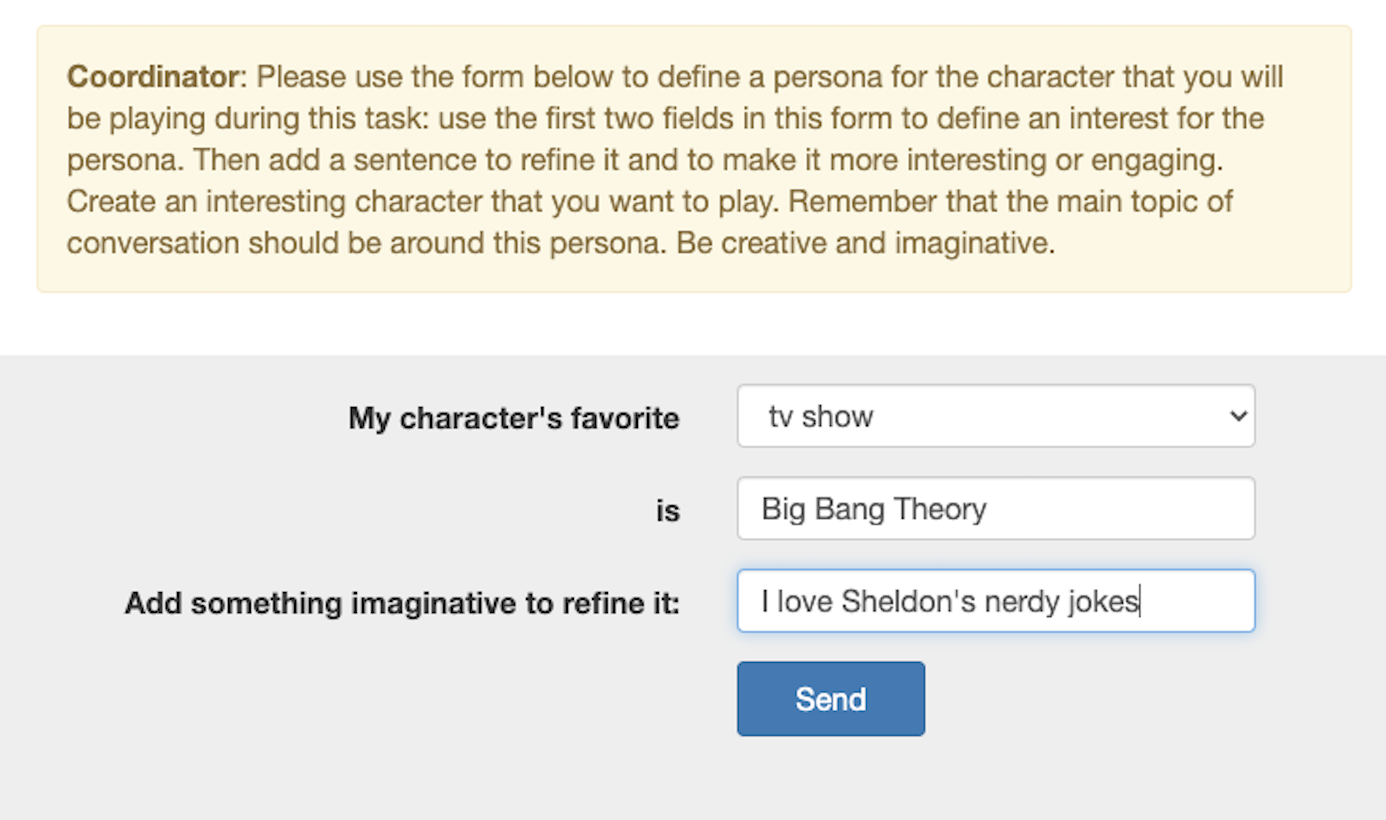}
    \caption{Crowdworker persona entry screenshot.}
    \label{fig:persona_collection}
\end{figure}

For the topics-based setting, we selected 7 general topics: 
(1) fashion (brand, designer or clothing type), (2) books (book, author), (3) music (artist, band, song, singer), (4) movies/TV (TV show, movie, actor, director), (5) sports (team, athlete), (6) hobby/game, (7) item to buy/recently bought. For a given crowdworker, we pick two of these topics at random for them to choose between. Then they fill in the following sentence ``My character's favorite <chosen\_topic\_area> is <specific\_item>'' and also write another imaginative sentence to refine it further. E.g. ``My favorite TV show is Big Bang Theory'' and ``I love Sheldon's nerdy jokes''.  See the screenshot example in \autoref{fig:persona_collection}.
This helps guarantee our conversations in the dataset are diverse and about a wide variety of topics and entities.

\section{Further Experimental Details}

\subsection{Model Training Details}

The majority of the models trained in the paper (using BART-Large), with retrieval augmentation, were trained on 4 32-GB GPUs, using the Adam \cite{kingma2014adam} optimizer, sweeping over learning rates between 1e-6 and 5e-5. During training, we used a batchsize of 16 and a linear LR scheduler with 100 warmup updates. We perform early stopping based on model perplexity evaluated on the validation set.

We retrieved $N = 5$ documents for each example. When using FAISS-based methods, the documents were given to the model in 100-word chunks. When using search engine-based methods, the first 256 tokens (according to the model's dictionary) of each document were given to the model.

\begin{figure}[t!]
  %   \centering
  \includegraphics[width=7.9cm]{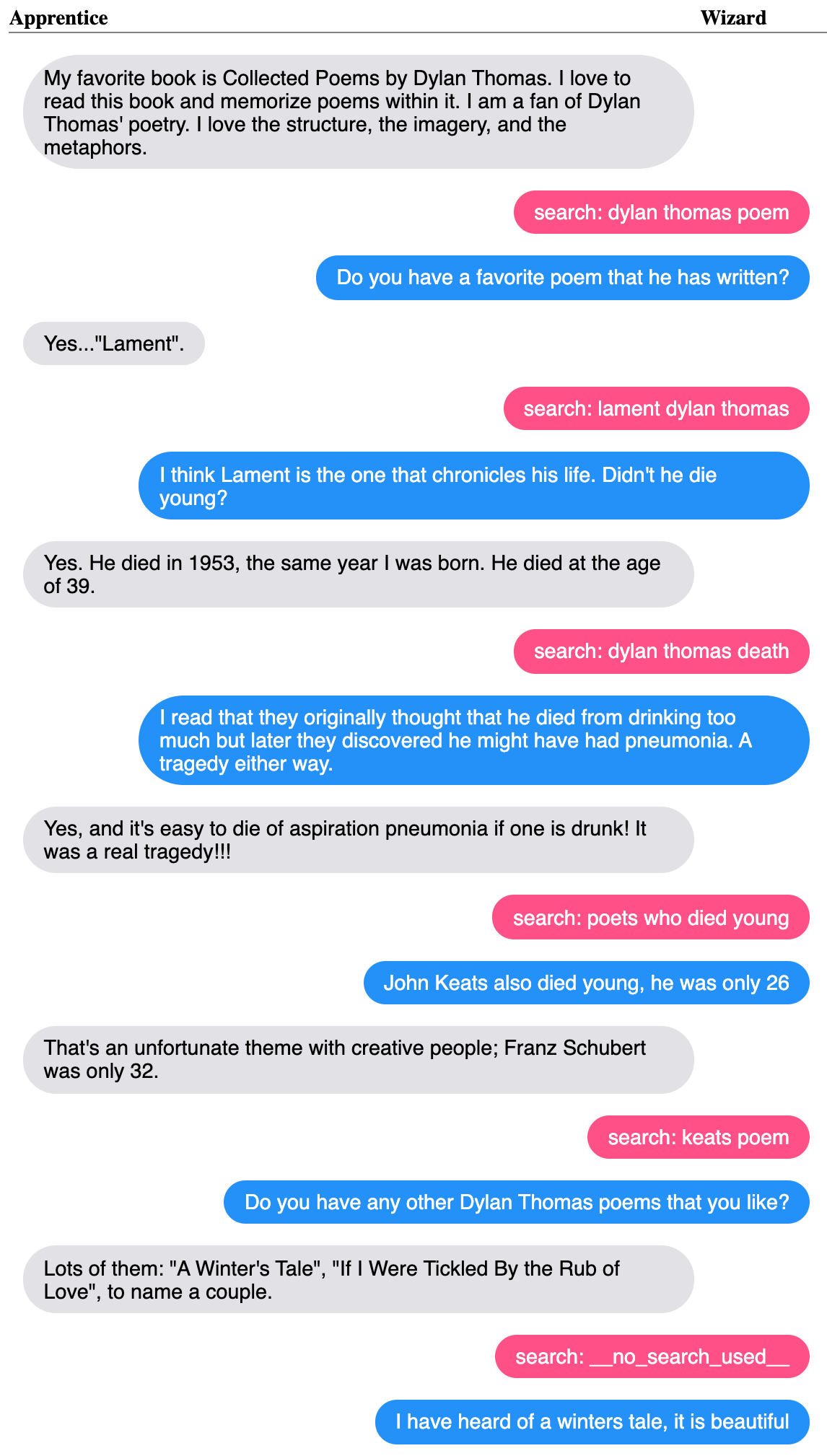}
    \caption{Additional example human-human conversation from the Wizard of the Internet training set. The role of the Wizard on the right-hand side involves performing internet searches, and then writing appropriate responses to the Apprentice given the viewed web documents (not shown).}
    \label{fig:train_example2}
\end{figure}

\subsection{Search Query Generation}

\subsubsection{Training Details}
Our search query generators are BART-Large models trained to produce human search queries given the dialogue context. The models were trained on 4 32-GB GPUs, using the Adam \cite{kingma2014adam} optimizer with a learning rate of 1e-5, batchsize of 64, and a linear LR scheduler with 100 warmup updates. We perform early stopping based on model perplexity evaluated on the validation set.

\subsubsection{Query Generation Performance}

To evaluate the performance of our search query generators, we take a look at some downstream metrics; that is, not only do we measure generation metrics on the query generation task, but also measure how good the search results are. Suppose we have the following three sets for each wizard search in the dataset: 1) $R = \{r_1, r_2, ..., r_k\}$, the set of \textbf{gold} retrieved documents; 2) $D = \{d_1, ..., d_m\}$, the set of documents \textbf{selected} by the wizard when conditioning their response; and 3) $S = \{s_1, ..., s_k\}$, the set of search results with the \textbf{generated} search query. We consider the following three metrics:

\begin{itemize}
    \item \% in Top 5: The percentage of all $r_i$ that are present in $S$.
    \item Average F1: For each $s_i$, compute the F1 word overlap with respect to all $r_i$ and determine the maximum F1 score; then, take the average of these max scores over all $s_i$.
    \item Gold Recall at 5: The proportion of the time any $d_i$ is in $S$.
\end{itemize}

We show results in \autoref{tab:sq_downstream} for two decoding schemes for our query generation models. The most important to note is that we obtain the gold document nearly 25\% of the time.

\begin{table}[h]
    \small
    \center
    \begin{tabular}{ll|rrr}
    Beam & Min Beam \\
    Size & Length & \% Top 5 & Avg. F1 & Gold R@5 \\
    \hline
    1 & 1 & 17.2 & 38.9 & 24.6 \\
    3 & 3 & 16.8 & 39.0 & 24.9 \\
    \end{tabular}
    \caption{{\bf Downstream retrieval performance of search query generators}.}
    \label{tab:sq_downstream}
\end{table}

\subsubsection{Effects of Decoding Algorithm}
We evaluated the effect of beam size and minimum beam length in search query generation. One may hypothesize that having a longer and more refined search query increases the chance of retrieving better documents, which might improve the overall performance of models that rely on search engines. However,  we observe little change in automatic metrics when changing these hyperparameters, see  \autoref{tab:krr}.

\begin{table}[h]
    \small
    \center
    \begin{tabular}{ll|rrr}
    Beam size & Min beam length & PPL & F1 & KF1 \\
    \hline
    1 & 1 & 16.4 & 17.9 & 6.9 \\
    3 & 3 & 16.4 & 17.8 & 6.9 \\
    3 & 4 & 16.5 & 17.9 & 6.8 \\
    \end{tabular}
    \caption{{\bf Effect of beam size and minimum beam length during search query generation}. Search engine FiD (CC+Wikipedia).}
    \label{tab:sqbeam}
\end{table}

\subsection{WoW Baselines}

We note that several of the WoW-trained baselines utilize a "search query" setup. The search query generators for these models were not trained on the WizInt dataset, but rather were trained to generate the title of the Wikipedia page corresponding to the gold selected knowledge in the WoW dataset.

\begin{figure*}[h!]
    \centering
    \center\includegraphics[width=15cm]{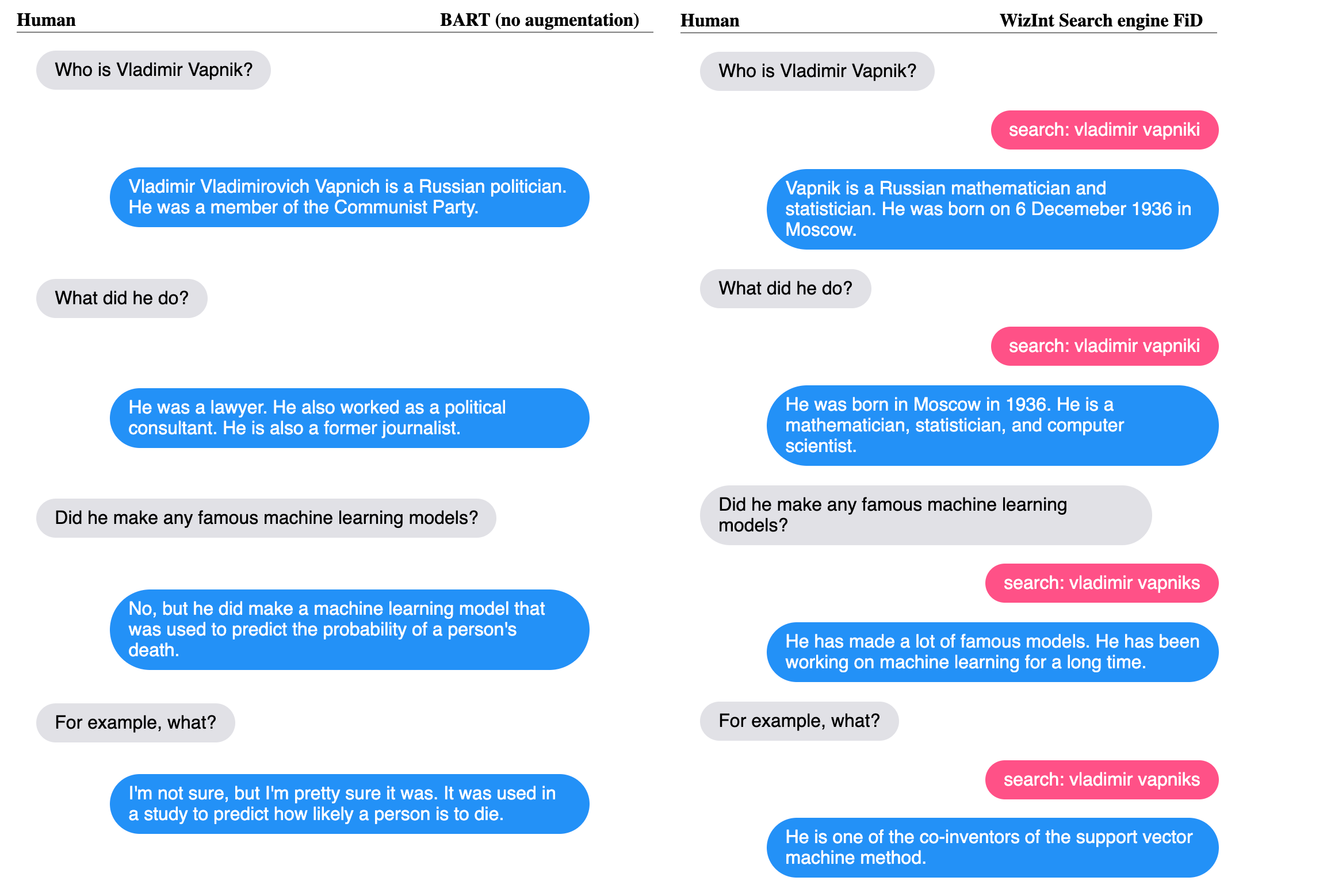}
    \caption{Cherry picked example comparing a nonaugmented model  (BART trained on WizInt, left) to a model with internet-augmentation (right). The right model is able to correctly describe attributes of Vladimir Vapnik, whereas the left model hallucinates details.}
    \label{fig:cherry4}
\end{figure*}

\end{document}